\documentclass[runningheads]{llncs}

\usepackage[T1]{fontenc}
\usepackage{amsmath}
\usepackage{amssymb}
\usepackage{cite}
\usepackage{xspace}
\usepackage[hyphens]{url}
\usepackage[dvipsnames]{xcolor}
\definecolor{linkblue}{rgb}{0,0,1.0}
\usepackage[labelfont=bf,font=small,tableposition=bottom]{caption}
\usepackage[skip=3pt]{subcaption}

\makeatletter
\DeclareRobustCommand\onedot{\futurelet\@let@token\@onedot}
\def\@onedot{\ifx\@let@token.\else.\null\fi\xspace}

\makeatother

\usepackage{bbm}
\usepackage{graphicx}
\usepackage{booktabs}
\usepackage[accsupp]{axessibility}
\usepackage[breaklinks,colorlinks,citecolor=linkblue]{hyperref}
\usepackage[capitalize]{cleveref}
\crefname{section}{Sec.}{Secs.}
\Crefname{section}{Section}{Sections}
\crefname{table}{Tab.}{Tabs.}
\Crefname{table}{Table}{Tables}

\begin{document}

\title{Readout $\neq$ Recovery: Dissociating Coordinate Emission from Visual-Corruption Repair in Vision-Language Models} 

\titlerunning{Readout $\neq$ Recovery}

\author{Drandreb Earl Juanico}

\authorrunning{D.~E.~Juanico}

\institute{}

\maketitle

\begin{abstract}
VLM bounding-box localization is both language generation and spatial commitment. Parseable fields such as \texttt{bbox\_2d} make localization easy to score, but dimensions that emit coordinate tokens need not repair localization after visual evidence is damaged. We study this readout/recovery separation in Qwen3-VL-4B-Instruct on single-object COCO grounding. We compare clean coordinate-token readout rankings with corruption-derived repair rankings, using object-mask endpoint replacement for recovery and clean-input flooring for depth localization. In Qwen3-VL, coordinate-token rankings are inert through layer 24, load-bearing from layers 32--35, and peak at layer 34; corruption-derived rankings harm layers 16--24 but become beneficial near layer 35/final. A Kimi-VL-A3B diagnostic shows a matching output-proximal transition despite a different box format. Object-mask recovery separates rank budgets: $k=250$ shows necessity, $k=500$ shows Top-$k$ restoration above random, and $k=d/2$ is largely capacity-driven. Partial-occlusion sweeps reveal that high-overlap coordinate-token sets can hurt at $k=1000$ and help mainly at half-width, while population corruption-derived sets provide no reliable fixed repair set. Edge-attribution patching shows coordinate-token paths are high precision but low recall for detection recovery, and RMSNorm quasi-layer controls do not close the endpoint-repair gap. Endpoint coordinate triage is therefore a useful circuit prior, but occlusion recovery requires a separate benchmark.
\end{abstract}

\section{Introduction}
\label{sec:introduction}

Modern VLMs increasingly express localization through language-like structured outputs. In object-detection-as-generation systems, boxes and labels can be serialized as token sequences; in grounded multimodal LLMs, object references can be linked to coordinate or location tokens; and in instruction-tuned VLMs, a response may contain a parseable field such as \texttt{bbox\_2d} followed by numeric coordinates \cite{chen2021pix2seq,peng2023kosmos2,chen2023shikra,you2024ferret,bai2025qwen3vl}. This makes grounding measurable: a generated box can be parsed, projected into image coordinates, and scored against annotated objects with COCO-style localization metrics \cite{lin2014microsoft}. It also creates a mechanistic ambiguity. The model must \emph{say} a box, but saying a box is not identical to preserving the visual evidence that makes the box correct.

This paper studies that ambiguity as a circuit-tracing problem. Endpoint-based mechanistic work often begins with a behavior, defines a metric, and then corrupts or patches activations to identify components that preserve or restore the endpoint \cite{meng2022locating,wang2022interpretability,conmy2023towards}. For VLM grounding, the final hidden state before the LM head is a natural endpoint because it is the last internal representation from which the localization tokens are emitted. A dimension that contributes to the logit of \texttt{bbox\_2d} or to a coordinate token is therefore a plausible readout-facing reference. The open question is whether that readout reference is also a repair-bearing substrate when the object evidence in the image is damaged.

We separate three ranking signals. The first, \texttt{bbox\_2d} readout triage, is the motivation: the field token is stable across examples and provides a convenient endpoint for identifying dimensions that help enter the structured localization format. The second, coordinate-token triage, is the main clean endpoint ranking: it scores final-site dimensions by their contribution to the actual numeric coordinate tokens emitted for each sample. This ranking is closer to spatial content than the field token, although it is still derived from clean readout behavior. The third, corruption-derived ranking, is repair-aware: it aggregates dimensions that repeatedly align with clean-to-object-mask repair displacements across the ranking population. The first two rankings ask about output-interface leverage; the third asks about object-mask repair leverage.

Object masking supplies the primary counterfactual for the repair question. We replace the ground-truth object region with a mean-color patch, damaging local evidence for the queried object while preserving the prompt and much of the surrounding scene. We also include partial occlusion as a softer corruption: a constant-valued rectangular or elliptical patch covers only a fraction of the object box, leaving part of the object visible. This is a practically motivated extension because many real occlusion failures involve partial visibility rather than complete erasure, although the synthetic patch remains an occlusion-like probe rather than a full naturalistic occluder. These corruptions are not intended as full naturalistic occlusion benchmarks. They are occlusion-like mechanistic probes: after object evidence is degraded to different degrees, do selected endpoint dimensions from the clean trace recover localization behavior more efficiently than arbitrary dimensions, and does clean coordinate-token triage become more useful when some object evidence remains?

Object-mask endpoint recovery uses clean endpoint replacement. The image is first object-masked and the model is run on the corrupted input. At the endpoint, selected dimensions in the corrupted hidden state are then replaced by the corresponding clean values from the uncorrupted-image trace. Recovery therefore tests whether the chosen endpoint dimensions are sufficient to transfer clean localization information into the corrupted run. Elementwise flooring is reserved for clean-input depth triage, where no upstream corruption is present and the goal is to determine where a fixed ranking becomes load-bearing across decoder depths.

In the primary Qwen3-VL-4B-Instruct experiments, the results support a readout/recovery separation rather than a single global ranking. Coordinate-token rankings are inert through layer 24, load-bearing from layers 32--35, and null at the final readout site. Corruption-derived rankings are also non-monotonic: the same half-width set (Top-1280 of 2560 hidden dimensions) that helps near layer 35 (the penultimate layer for this architecture) and the final site is harmful across layers 16--24. Endpoint recovery further shows a rank-budget dissociation, where rank budget means the number $k$ of endpoint dimensions replaced by clean values. For coordinate-token rankings under object masking, $k=250$ shows necessity without stand-alone restoration, $k=500$ shows a Top-$k$ restorative effect above random, and $k=1280=d/2$ is largely capacity-driven. The population corruption-derived curve is asymmetric: exclusion costs appear already at $k=250$, inclusion evidence emerges at $k=500$, and half-width recovery is capacity-dominated. Thus, clean coordinate-token triage identifies a meaningful coordinate-generation subspace, whereas corruption-derived rankings remain the direct benchmark for hard object-mask repair. A targeted Kimi-VL-A3B-Instruct check strengthens the model-generality of the coordinate-token result: despite emitting normalized bracketed boxes rather than \texttt{bbox\_2d} JSON, Kimi's coordinate-token-ranked half-width subspace is inert or mildly helpful at observed early-to-mid sites, begins to separate at layer 24, and becomes selectively load-bearing at layer 26 and final. Partial occlusion then refines the practical message: when object evidence is degraded but not erased, high-overlap coordinate-token index sets do not automatically become useful recovery sets; under partial occlusion they are sub-random at $k=1000$ and become helpful mainly at half-width. An edge-attribution patching diagnostic under the same partial-occlusion setting then asks where the readout-targeted gap is routed: the top readout-targeted edges are most concentrated at small ablation budgets, but detection-critical paths extend beyond that subspace. Finally, a RMSNorm quasi-layer control shows that transporting \texttt{bbox\_2d} endpoint directions through the final RMSNorm Jacobian does not make endpoint rankings substantially more repair-like, ruling out final normalization as the main source of the readout/recovery gap.

Our contributions are threefold. First, we formulate VLM grounding interpretability as a distinction between coordinate-emission geometry and visual-corruption recovery geometry. Second, we define object-mask repair as clean endpoint replacement, while reserving flooring for clean-input depth localization. Third, we report main-text evidence that coordinate-token and corruption-derived rankings have distinct, non-monotonic depth profiles, add a targeted Kimi-VL-A3B coordinate-token depth-triage check, show that partial occlusion separates cross-split index-set overlap from behavioral recovery, and add a RMSNorm control with explicit normalization showing that the dissociation is not removed by transporting endpoint attribution across the final normalization boundary.

\section{Related Work}
\label{sec:related-work}

\paragraph{Structured localization in VLMs.}
Vision-language systems have increasingly moved from implicit region-language alignment toward explicit localization in generated language. Pix2Seq casts object detection as sequence generation \cite{chen2021pix2seq}; Kosmos-2 represents grounded spans with bounding boxes and location tokens \cite{peng2023kosmos2}; Shikra supports spatial coordinate inputs and outputs in natural language \cite{chen2023shikra}; and Ferret combines discrete coordinates with continuous region features for referring and grounding at variable granularity \cite{you2024ferret}. Qwen3-VL continues this trend in an instruction-tuned VLM family with explicit spatial reasoning capabilities \cite{bai2025qwen3vl}. Kimi-VL provides a complementary efficient VLM setting with a different localization output format, which we use only as a targeted cross-model check rather than as the primary experimental platform \cite{team2025kimi}. We use this structured-output setting to ask whether dimensions that support coordinate emission also support repair when object evidence is damaged.

\paragraph{Endpoint tracing and causal intervention.}
Mechanistic interpretability commonly starts from a behavioral endpoint and asks which internal components causally affect it. Causal mediation and causal tracing intervene on hidden states or mediators to identify components implicated in a prediction \cite{vig2020investigating,meng2022locating}. Path patching and the IOI circuit study use clean/corrupt distributions and causal interventions to trace a language behavior through residual-stream components \cite{wang2022interpretability}. ACDC systematizes this endpoint-driven workflow by choosing a dataset and metric, then pruning the computational graph to preserve behavior \cite{conmy2023towards}. Our study follows the endpoint tradition but adds a transfer test: a localization endpoint is useful for tracing only if it is separately evaluated against visual-corruption repair.

\paragraph{Faithfulness, corruption, and normalization.}
Attribution-like signals can identify output-facing sensitivity without proving that the highlighted components implement the behavior \cite{jain2019attention,serrano2019attention,jacovi2020towards}. Visual corruptions and occlusion similarly reveal that recognition can depend on evidence not captured by a clean endpoint alone \cite{hendrycks2019benchmarking,michaelis2019benchmarking,wang2020robust}. Finally, RMSNorm and LayerNorm compute shared sample-wise normalizers across dimensions \cite{ba2016layer,zhang2019rmsnorm}, which can complicate dimension-level interventions. We therefore include a quasi-layer RMSNorm control in the broader evaluation suite, but the readout/recovery account does not rely on RMSNorm as the explanation for the separation.

\section{Method}
\label{sec:method}

\paragraph{Model, data, and endpoint.}
We study Qwen3-VL-4B-Instruct on single-object grounding prompts from COCO val2017. The text decoder has hidden width $d=2560$ and 36 decoder layers in the released 4B configuration \cite{bai2025qwen3vl,qwen3vl4bconfig}. Each prompt names one object category and asks the model to localize it with a prompt of the form \texttt{Detect <object>. Provide correct bounding boxes.} The model is expected to return a structured localization response, e.g., \texttt{\{"bbox\_2d": [357, 216, 708, 549], "label": "spoon"\}}. We parse the generated \texttt{bbox\_2d} coordinate list and evaluate localization by $R_{50}$, recall at IoU $\geq 0.50$, using one-to-one bipartite matching between predicted and ground-truth boxes. Parseability is reported separately, unparseable outputs contribute zero recall, and Supp. Table~\ref{tab:parseability} decomposes representative effects into parseability and conditional localization (Supp. Methods~\ref{app:parseability}).

The primary endpoint is the final readout site: the output of the final RMSNorm immediately before the LM head. Decoder-depth analyses hook decoder-block outputs, with $\ell$ reserved for decoder-layer indices. All interventions are decode-only: they fire during autoregressive generation of localization-related tokens and do not alter the image/prompt prefill pass (Supp. Methods~\ref{app:sites-targets}).

\paragraph{Cross-model Kimi-VL diagnostic.}
We also run a targeted clean-input depth-triage diagnostic on Kimi-VL-A3B to test whether coordinate-token triage is tied to Qwen3-VL's output schema. Kimi-VL emits plain normalized boxes of the form \texttt{[x1, y1, x2, y2]} rather than a \texttt{bbox\_2d} JSON object. We therefore construct a Kimi-specific coordinate-token ranking directly from the LM head and generated coordinate tokens, then floor the Top-$1024$ dimensions, i.e., half of Kimi's $d=2048$ hidden width, at proportional decoder-depth sites and the final norm output (Supp. Methods~\ref{app:kimivl-coordinate-triage}). This diagnostic is a cross-model readout-triage check, not a second full recovery benchmark.

\paragraph{Readout directions and rankings.}
Let $i$ index samples, $t$ localization-related decode positions, $j$ hidden dimensions, and $q$ vocabulary tokens. Let $h_{i,t}\in\mathbb{R}^{d}$ be the hidden state and $W_q\in\mathbb{R}^{d}$ the LM-head vector for token $q$. For target token $y_{i,t}$, we use a contrastive readout direction
\begin{equation}
w_{i,t}=W_{y_{i,t}}-\overline W^{-}_{i,t},
\end{equation}
where $\overline W^{-}_{i,t}$ averages negative alternatives for the same output position (Supp. Methods~\ref{app:readout-directions}). Readout-based rankings use the per-dimension score $\Delta h_{i,t,j}w_{i,t,j}$, with $\Delta h_{i,t,j}\allowbreak=F_{\theta}(h^{\mathrm{clean}}_{i,t,j})\allowbreak-h^{\mathrm{clean}}_{i,t,j}$ and $\theta=-0.5$ (Eq.~\ref{eq:flooring}; Supp. Methods~\ref{app:flooring-motivation}). The \texttt{bbox\_2d} ranking aggregates this score at field-token positions; the coordinate-token ranking aggregates it at numeric coordinate-token positions. Because coordinate-token rankings depend on the negative-token pool, Supp. Table~\ref{tab:negpool-sensitivity} reports alternative-pool overlaps; we treat the chosen pool as part of the ranking definition and ground claims in held-out interventions rather than exact dimension identity. Corruption-derived rankings are different: they start from per-sample clean-to-object-mask endpoint displacement orderings, and the population corruption-derived ranking averages the resulting disruption scores across ranking samples before freezing it for evaluation. In local-set recovery, a fixed global set reuses one Top-$k$ set for every test sample; a sample-local readout set is recomputed from the clean trace; and a sample-specific corruption-derived set is recomputed from the clean--corrupt pair. Thus fixed population rows are out-of-sample tests of a population-level repair ranking, not clean-readout rankings or per-sample oracle rankings (Supp. Methods~\ref{app:corruption-derived-ranking};~\ref{app:local-transfer}).

\paragraph{Clean-input depth triage.}
Depth triage does not use object-mask corruption. It uses an elementwise final-layer probing intervention motivated by prior anonymized work~\cite{juanico2026flip}: the clean image/prompt input is run and selected hidden dimensions are floored at a specified decoder layer or the final site (Supp. Methods~\ref{app:flooring-motivation}):
\begin{equation}
\label{eq:flooring}
F_{\theta}(h_j)=\max(h_j,\theta),\qquad \theta=-0.5,
\end{equation}
with all unselected dimensions left unchanged. Flooring is an elementwise clamp, not zero ablation, quantile clipping, or batch normalization: coordinates above the threshold are unchanged, and only below-threshold coordinates are lifted to the fixed floor. We use $\theta=-0.5$ as a non-collapsing final-site operating point motivated by the final-layer probing intervention; we do not claim that this threshold is layer-calibrated. When raw-layer full flooring collapses performance, we therefore interpret that cell as a site-calibration warning rather than as evidence of rank selectivity. In this paper, flooring is used for clean-input depth localization; visual-corruption recovery is evaluated separately with clean endpoint replacement. For a fixed rank size $k$, we compare Top-$k$, complement, random-$k$, complement-random, and full flooring conditions. A negative top-random $z$ means the ranked Top-$k$ set is more damaging to floor than a size-matched random set; a positive value means it is less damaging or more beneficial than random. These clean-input experiments localize where a ranking is load-bearing, but they are not recovery experiments.

\begin{figure}[t]
\centering
\begin{minipage}{0.48\linewidth}
    \centering
    \includegraphics[width=\linewidth,trim=120 180 180 70,clip]{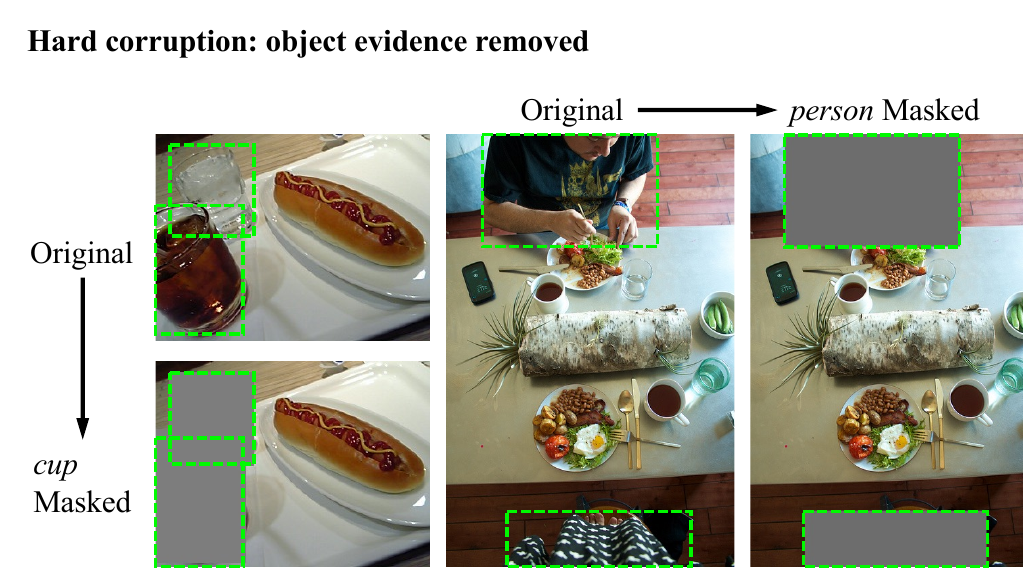}
    \vspace{-1mm}
    \textbf{(a) Object mask}
\end{minipage}
\hfill
\begin{minipage}{0.48\linewidth}
    \centering
    \includegraphics[width=\linewidth,trim=120 180 180 70,clip]{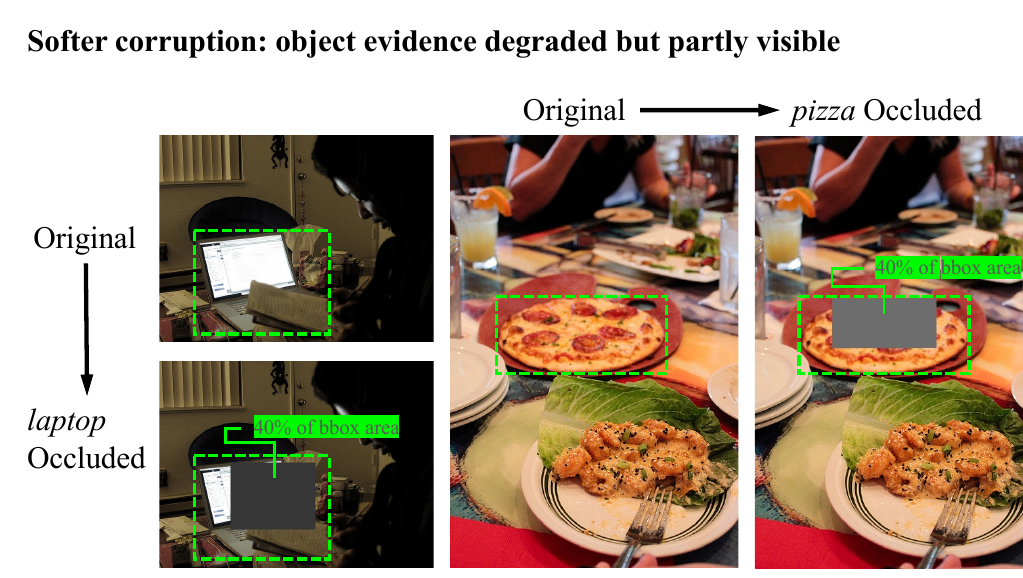}
    \vspace{-1mm}
    \textbf{(b) Partial occlusion}
\end{minipage}
\vspace{-1mm}
\caption{
Illustration of the two corruption settings. Object mask is the hard corruption: the queried object region is blanked, removing most local evidence for the target. Partial occlusion is the softer corruption: a patch is sampled inside the ground-truth box, leaving part of the queried object visible. This distinction motivates treating object mask as the hard repair benchmark and partial occlusion as the more practical occlusion-like sensitivity setting.
}
\label{fig:corruption-settings}
\end{figure}

\paragraph{Endpoint recovery under visual corruptions.}
Recovery experiments use a different operator. The image is first corrupted either by masking the full ground-truth object region with the per-image mean color or by applying partial occlusion inside the ground-truth box. For partial occlusion, the patch area is a fixed fraction of the box area, its position is sampled inside the box, and the rest of the object remains visible (Supp. Methods~\ref{app:partial-occlusion}). We use object masking as the hard repair benchmark and partial occlusion as the practical softer-corruption extension (Figure~\ref{fig:corruption-settings}), designed to test whether coordinate-token readout remains a useful transferable index-set prior when the target is degraded but not removed. The model is then run on the corrupted input. At the endpoint and localization-related decode positions, selected dimensions in the corrupted hidden state are replaced by their corresponding clean values:
\begin{equation}
h^{\mathrm{patch}}_{t,I}=h^{\mathrm{clean}}_{t,I},\qquad
h^{\mathrm{patch}}_{t,\bar I}=h^{\mathrm{mask}}_{t,\bar I}.
\end{equation}
Here $I$ is a hidden-dimension set (Supp. Methods~\ref{app:dimension-sets}), not a token-position set. This clean endpoint replacement is the recovery intervention used for $k$-scaling endpoint recovery and local-set endpoint recovery tests. It asks whether a selected subset transfers clean localization information into the corrupted run. Object mask is the harder primary corruption; partial occlusion is the softer occlusion-like extension in which residual visible-object evidence can make clean coordinate-token readout more informative as an index-set prior. Full endpoint recovery replaces all endpoint dimensions and serves as the recovery ceiling.

\paragraph{Evaluation suite.}
We use five complementary tests. First, layer-resolved selective flooring moves the same ranked dimension sets across decoder depths and final readout to determine where a ranking is load-bearing. Second, $k$-scaling endpoint recovery replaces Top-$k$, Bottom-$k$, complement, random-$k$, complement-random, and full dimension sets with their clean endpoint values to separate ranking-enriched repair from generic dimensional capacity, using object mask as the primary corruption and partial occlusion as a softer sensitivity check. Third, local-set endpoint recovery compares fixed global, split-local, sample-local readout, sample-specific corruption-derived, bottom-ranked, and random index sets under the same paired-corruption convention. The partial-occlusion version is read as a practical transfer test for coordinate-token triage rather than as a weaker copy of the object-mask benchmark. In all local-set tables, fixed global set means one fixed Top-$k$ set reused for every test sample; split-local set is fixed per split; sample-local readout set is recomputed from the current clean sample trace; sample-specific corruption-derived set is recomputed from the current clean--corrupt sample pair; and bottom-ranked set is the Bottom-$k$ set from the corresponding fixed ranking (Supp. Methods~\ref{app:local-transfer}). Fourth, an edge-attribution patching diagnostic scores attention and MLP edges under partial occlusion against a coordinate-token readout-ranked Top-$k$ objective, its complement, and a random-$k$ endpoint objective, then validates the resulting edge rankings by zero-ablation (Supp. Methods~\ref{app:eap-partial-occlusion}). Fifth, a quasi-layer RMSNorm control maps post-norm endpoint and repair directions through the RMSNorm Jacobian to test whether final normalization alone manufactures the readout/recovery dissociation (Supp. Methods~\ref{app:layer-resolved-flooring}--\ref{app:rmsnorm-control}). 

\section{Results}
\label{sec:results}

This section summarizes the submitted evaluation. Single-split rows are effect-size diagnostics. Tables~\ref{tab:coord-depth}--\ref{tab:corruption-derived-depth} report clean-input depth triage; Tables~\ref{tab:kscaling-recovery}--\ref{tab:local-set-recovery} report object-mask endpoint recovery; Table~\ref{tab:partial-occlusion} reports the partial-occlusion sensitivity check; and Table~\ref{tab:rmsnorm-main} reports the RMSNorm quasi-layer control. $z_{\mathrm{top}}$ compares Top-$k$ with random-$k$, and $z_{\mathrm{comp}}$ compares complement with complement-random. Supp. Table~\ref{tab:eap-ablation} and Supp. Fig.~\ref{fig:eap-heatmaps} report the partial-occlusion edge-attribution patching diagnostic. Supp. Tables~\ref{tab:parseability} and~\ref{tab:negpool-sensitivity} report parseability and negative-pool sensitivity controls.

\begin{table}[t]
\centering
\caption{Clean-input depth triage on Qwen3-VL using coordinate-token rankings. The Qwen Top-1280 set is null or protective through layer 17, complement-load-bearing at layers 21--24, and top-load-bearing from layers 32--35, with peak specificity at layer 34. The schema mirrors Table~\ref{tab:kimivl-depth}.}
\label{tab:coord-depth}
\scriptsize
\setlength{\tabcolsep}{3pt}
\resizebox{\textwidth}{!}{%
\begin{tabular}{llrrrrrrl}
\toprule
Layer & $n$ & Base & $\Delta$Full & $\Delta$Top & $\Delta$Comp & $z_{\mathrm{top}}$ & $z_{\mathrm{comp}}$ & Interpretation \\
\midrule
$8$ & $3741$ & $.5871$ & $-.0014$ & $+.0022$ & $-.0002$ & $+1.34$ & $-.07$ & null; controls near baseline \\
$16$ & $3741$ & $.5871$ & $-.0595$ & $+.0014$ & $-.0112$ & $+.41$ & $\mathbf{-2.19}$ & complement harmful \\
$17$ & $3741$ & $.5871$ & $-.0435$ & $+.0036$ & $-.0061$ & $+.20$ & $-1.95$ & mild complement harm \\
$21$ & $3741$ & $.5871$ & $\mathbf{-.2243}$ & $+.0004$ & $\mathbf{-.0540}$ & $+.97$ & $\mathbf{-3.70}$ & complement-load-bearing \\
$24$ & $2997$ & $.5904$ & $\mathbf{-.2692}$ & $+.0101$ & $\mathbf{-.0368}$ & $+.87$ & $\mathbf{-3.98}$ & complement-load-bearing \\
$32$ & $2984$ & $.5899$ & $\mathbf{-.4603}$ & $\mathbf{-.0340}$ & $+.0006$ & $\mathbf{-11.84}$ & $-.83$ & top load-bearing; full collapse \\
$33$ & $3003$ & $.5871$ & $-.0293$ & $\mathbf{-.0691}$ & $\mathbf{-.0252}$ & $\mathbf{-20.72}$ & $\mathbf{-17.26}$ & broad sensitivity; top worse \\
$34$ & $3006$ & $.5871$ & $-.0044$ & $\mathbf{-.1722}$ & $\mathbf{-.0235}$ & $\mathbf{-40.79}$ & $\mathbf{-2.59}$ & peak top specificity \\
$35$ & $2974$ & $.5841$ & $+.0010$ & $-.0051$ & $-.0039$ & $\mathbf{-10.31}$ & $-1.57$ & weak top-specific effect \\
final & $3741$ & $.5871$ & $+.0021$ & $+.0029$ & $-.0009$ & $+1.04$ & $-.69$ & final-site null \\
\bottomrule
\end{tabular}}
\end{table}

\paragraph{Coordinate-token rankings separate whole-layer fragility from Top-$k$ selectivity.}
Table~\ref{tab:coord-depth} separates broad layer fragility from Top-$k$ selectivity. Full flooring severely damages layers 21, 24, and 32, but Top-vs-random statistics identify which half of the hidden state carries the effect. The final site, layer 8, and layers 16--17 are null or weak for Top-1280. Layers 21--24 form a complement-load-bearing middle band: Top flooring is near random ($z_{\mathrm{top}}=+.97$ and $+.87$), while complement flooring is strongly harmful ($z_{\mathrm{comp}}=-3.70$ and $-3.98$). The coordinate-token Top-1280 set becomes specifically load-bearing from layer 32 onward, with broad sensitivity at layer 33 and peak top specificity at layer 34. Thus, the clean coordinate-token subspace is not uniformly active; it shifts from complement-dominated mid-depth fragility to top-specific late load-bearing.

\begin{table}[t]
\centering
\caption{Clean-input depth triage on Kimi-VL-A3B using coordinate-token rankings. The Kimi Top-1024 set is the half-width analogue of Qwen's Top-1280 set. The schema mirrors Table~\ref{tab:coord-depth}; entries are split-mean values.}
\label{tab:kimivl-depth}
\scriptsize
\setlength{\tabcolsep}{3pt}
\resizebox{\textwidth}{!}{%
\begin{tabular}{llrrrrrrl}
\toprule
Layer & $n$ & Base & $\Delta$Full & $\Delta$Top & $\Delta$Comp & $z_{\mathrm{top}}$ & $z_{\mathrm{comp}}$ & Interpretation \\
\midrule
$6$     & $1502$ & $.210$ & $+.006$ & $+.002$ & $+.003$ & $-.75$ & $+.75$ & early null \\
$12$    & $1502$ & $.210$ & $+.006$ & $+.008$ & $+.004$ & $+1.51$ & $+.12$ & mild/nonselective benefit \\
$13$    & $1502$ & $.210$ & $+.004$ & $+.005$ & $+.003$ & $+.70$ & $-.44$ & early null \\
$16$    & $1502$ & $.210$ & $-.001$ & $+.005$ & $+.007$ & $+.56$ & $+.88$ & nonselective helpful \\
$18$    & $1502$ & $.210$ & $+.011$ & $+.007$ & $+.004$ & $+1.18$ & $-.25$ & early null \\
$24$    & $1502$ & $.210$ & $+.005$ & $-.008$ & $\mathbf{+.011}$ & $-.94$ & $\mathbf{+1.42}$ & transition onset \\
$25$    & $1502$ & $.210$ & $+.008$ & $-.006$ & $+.011$ & $-1.43$ & $+.11$ & transition; complement mixed \\
$26$    & $1502$ & $.210$ & $+.002$ & $\mathbf{-.022}$ & $\mathbf{+.015}$ & $\mathbf{-6.01}$ & $+1.32$ & top load-bearing \\
final   & $1502$ & $.210$ & $+.005$ & $\mathbf{-.031}$ & $\mathbf{+.014}$ & $\mathbf{-9.17}$ & $+.54$ & terminal top load-bearing \\
\bottomrule
\end{tabular}}
\end{table}

\paragraph{Kimi-VL shows an early-to-terminal coordinate-token transition.}
Table~\ref{tab:kimivl-depth} mirrors the Qwen depth-triage columns in Table~\ref{tab:coord-depth} so the two readout-depth diagnostics can be compared directly. Layers 6--18 remain null or mildly helpful: Top-$1024$ flooring stays close to random controls and parseability is flat. Layers 24--25 mark the transition. Top flooring is weakly negative, while complement flooring often helps more, indicating that complement noise appears before the coordinate-token Top half becomes irreplaceable; the mixed complement control at layer 25 is the only local exception. At layer 26 the pattern crystallizes in both splits, with a step change to large negative Top-vs-random statistics, followed by an even stronger final-site effect. The full-floor rows nearly cancel the Top penalty with the complement benefit, indicating a late functional polarity: coordinate-token Top-half dimensions carry signal, while the complement behaves partly like removable noise. The associated parseability drop is also depth-gated, emerging around layers 24--25 and growing to about $3$--$5$ percentage points at layer 26/final. Thus, Kimi-VL shows a single-step functional transition at the penultimate decoder layer, with organizational restructuring beginning slightly earlier.

\begin{table}[t]
\centering
\caption{Clean-input depth triage using population corruption-derived rankings. The Top-1280 set is null at layer 8, harmful across layers 16--24, transitional at layer 32, necessary before sufficient at layers 33--34, and dual-criterion load-bearing at layer 35; the final site retains a weaker sufficiency signal. $S$ is the number of held-out splits contributing to the row. Parse(full) reports parseability under full flooring. The selectivity gap is $\Delta$Top$-\Delta$Comp.}
\label{tab:corruption-derived-depth}
\scriptsize
\setlength{\tabcolsep}{2.5pt}
\resizebox{\textwidth}{!}{%
\begin{tabular}{llrrrrrrrrr}
\toprule
Layer & $S$ & $n$ & Base & $\Delta$Full & Parse(full) & $\Delta$Top & $z_{\mathrm{top}}$ & $\Delta$Comp & $z_{\mathrm{comp}}$ & Selectivity gap \\
\midrule
$8$ & $5$ & $3741$ & $.5871$ & $-.0014$ & $.974$ & $+.0001$ & $-1.26$ & $+.0010$ & $+.67$ & $-.0009$ \\
$16$ & $5$ & $3741$ & $.5871$ & $-.0595$ & $.881$ & $\mathbf{-.0731}$ & $\mathbf{-8.78}$ & $+.0019$ & $+.60$ & $\mathbf{-.0750}$ \\
$17$ & $5$ & $3741$ & $.5871$ & $-.0435$ & $.882$ & $\mathbf{-.0623}$ & $\mathbf{-8.02}$ & $-.0001$ & $-.28$ & $\mathbf{-.0623}$ \\
$21$ & $4$ & $2997$ & $.5904$ & $-.2279$ & $.587$ & $\mathbf{-.3412}$ & $\mathbf{-44.68}$ & $-.0015$ & $+.68$ & $\mathbf{-.3397}$ \\
$24$ & $3$ & $2240$ & $.5952$ & $-.2704$ & $.476$ & $\mathbf{-.2590}$ & $\mathbf{-20.69}$ & $+.0006$ & $+.88$ & $\mathbf{-.2597}$ \\
$32$ & $4$ & $2984$ & $.5899$ & $-.4603$ & $.216$ & $-.0043$ & $\mathbf{-2.19}$ & $+.0008$ & $-.56$ & $-.0051$ \\
$33$ & $4$ & $3003$ & $.5953$ & $-.0290$ & $.932$ & $+.0036$ & $+.53$ & $\mathbf{-.0016}$ & $\mathbf{-2.10}$ & $+.0053$ \\
$34$ & $3$ & $2239$ & $.5681$ & $+.0007$ & $.979$ & $+.0022$ & $+1.58$ & $\mathbf{-.0037}$ & $-1.92$ & $+.0059$ \\
$35$ & $4$ & $2974$ & $.5841$ & $+.0010$ & $.978$ & $\mathbf{+.0084}$ & $\mathbf{+17.22}$ & $\mathbf{-.0057}$ & $\mathbf{-2.27}$ & $\mathbf{+.0141}$ \\
final & $5$ & $3741$ & $.5871$ & $+.0017$ & $.977$ & $\mathbf{+.0050}$ & $\mathbf{+2.02}$ & $-.0013$ & $-.75$ & $+.0062$ \\
\bottomrule
\end{tabular}}
\end{table}

\paragraph{Population corruption-derived rankings show a separate multi-phase depth profile.}
Table~\ref{tab:corruption-derived-depth} uses the same clean-input flooring operator but a population corruption-derived ranking. Each site now has at least three held-out splits. Layer 8 is null. Layers 16--24 form a strong harmful band in which Top-1280 flooring is much worse than random, peaking at layer 21, while complement flooring remains near baseline. Layer 32 is transitional: full flooring is catastrophic and Parse(full) falls to $.216$, but Top-1280 flooring is only weakly negative and the complement is near random. Layers 33--34 show necessity before sufficiency: complement flooring becomes harmful or borderline harmful, but Top flooring is not yet beneficial. Layer 35 is the sole dual-criterion site, with Top flooring above random ($z_{\mathrm{top}}=+17.22$) and complement flooring below complement-random ($z_{\mathrm{comp}}=-2.27$). The final site keeps only a weaker sufficiency signal. Thus, the population corruption-derived set is dormant early, destructive when perturbed mid-computation, and most coherently load-bearing at the last decoder layer before final normalization.

\begin{table}[t]
\centering
\caption{$k$-scaling endpoint recovery under clean endpoint replacement. Entries are recovery percentages relative to the object-mask baseline and full-recovery ceiling; $\pm$ reports standard deviation across per-seed means where shown. $z_{\mathrm{top}}$ and $z_{\mathrm{comp}}$ compare Top-$k$ and complement against matched random controls. For $k=1280=d/2$, Bottom-$k$ and complement are identical.}
\label{tab:kscaling-recovery}
\scriptsize
\setlength{\tabcolsep}{2.5pt}
\resizebox{\textwidth}{!}{%
\begin{tabular}{llrrrrrrl}
\toprule
Ranking source & $k$ & Top & Random & Bottom & Complement & Comp-random & $z_{\mathrm{top}}$ / $z_{\mathrm{comp}}$ & Interpretation \\
\midrule
Coordinate-token & $250$  & $3.27\pm1.31$ & $2.97\pm.76$ & $2.02\pm.73$ & $\mathbf{90.31\pm.81}$ & $\mathbf{94.79\pm.96}$ & $+.92$ / $\mathbf{-5.60}$ & necessity only \\
Coordinate-token & $500$  & $\mathbf{10.60\pm2.05}$ & $\mathbf{8.77\pm1.67}$ & $6.00\pm1.57$ & $90.60\pm.61$ & $91.04\pm.64$ & $\mathbf{+2.35}$ / $-.37$ & Top-$k$ restoration emerges \\
Coordinate-token & $1280$ & $76.67\pm2.74$ & $73.73\pm2.41$ & $72.15\pm3.80$ & $72.15\pm3.80$ & $74.44\pm2.57$ & $+1.19$ / $-1.00$ & capacity-dominated \\
\addlinespace[1mm]
Corruption-derived & $250$  & $2.89\pm0.84$ & $2.93\pm0.72$ & $3.58\pm1.08$ & $\mathbf{89.01\pm0.87}$ & $\mathbf{94.31\pm0.96}$ & $-.17$ / $\mathbf{-8.41}$ & exclusion cost only \\
Corruption-derived & $500$  & $\mathbf{10.95\pm1.68}$ & $\mathbf{8.76\pm1.49}$ & $9.40\pm1.32$ & $\mathbf{88.64\pm0.77}$ & $\mathbf{91.04\pm0.57}$ & $\mathbf{+2.62}$ / $\mathbf{-2.09}$ & sufficiency and necessity \\
Corruption-derived & $1280$ & $69.80\pm2.75$ & $69.00\pm2.66$ & $68.29\pm4.27$ & $68.29\pm4.27$ & $69.76\pm2.45$ & $+.31$ / $-.79$ & capacity-dominated \\
\bottomrule
\end{tabular}}
\end{table}

\paragraph{$k$-scaling endpoint recovery separates necessity, restoration, and capacity.}
Table~\ref{tab:kscaling-recovery} shows the rank-budget logic. For coordinate-token rankings, $k=250$ shows necessity only: Top-$k$ restoration is random-like ($3.27\%$ versus $2.97\%$), but withholding them lowers complement recovery ($90.31\%$ versus $94.79\%$, $z=-5.60$). At $k=500$, Top-$k$ restoration beats random ($10.60\%$ versus $8.77\%$, $z=+2.35$) and the complement penalty vanishes. At $k=1280=d/2$, Bottom-$k$ equals complement and recovery is capacity-dominated. Population corruption-derived rankings show the stronger asymmetric signature: exclusion hurts at $k=250$, inclusion and exclusion both hold at $k=500$, and modes converge near $69$--$70\%$ at half-width. Coordinate-token recovery moves from necessity to sufficiency to capacity, whereas corruption-derived recovery directly targets hard-mask repair at $k=500$.

\begin{table}[t]
\centering
\caption{Object-mask endpoint recovery with local index sets under the primary clean endpoint-replacement operator. Entries report $R_{50}$; parentheses give the $z$ statistic against the matched random control. Fixed global means one Top-$k$ set reused for all test samples; split-local means one fixed set per split; sample-local readout is recomputed from the current clean test trace; sample-specific corruption is recomputed from the current clean--corrupt pair; bottom-ranked is the Bottom-$k$ set from the corresponding fixed ranking.}
\label{tab:local-set-recovery}
\scriptsize
\setlength{\tabcolsep}{2.5pt}
\resizebox{\textwidth}{!}{%
\begin{tabular}{llrrrrrr}
\toprule
Ranking source & $k$ & Random & Fixed global & Split-local range & Sample-local readout & Sample-specific corruption & Bottom-ranked \\
\midrule
Coordinate-token & $500$  & $.1200\pm.0035$ & $\mathbf{.1300\,(+2.87)}$ & $\mathbf{.1283\text{--}.1316\,(+2.38\text{--}+3.35)}$ & $.1236\,(+1.04)$ & $\mathbf{.1501\,(+8.66)}$ & $.1064\,(-3.91)$ \\
Coordinate-token & $1000$ & $.3542\pm.0600$ & $\mathbf{.2423\,(-1.86)}$ & $\mathbf{.2294\text{--}.2533\,(-2.08\text{--}-1.68)}$ & $.2480\,(-1.77)$ & $\mathbf{.4106\,(+0.94)}$ & $.3124\,(-0.69)$ \\
Coordinate-token & $1280$ & $.4547\pm.0122$ & $.4730\,(+1.49)$ & $.4651\text{--}.4777\,(+0.85\text{--}+1.88)$ & $.4671\,(+1.01)$ & $.4685\,(+1.13)$ & $.4483\,(-0.53)$ \\
\addlinespace[1mm]
Corruption-derived & $500$  & $.119\pm.0035$ & $\mathbf{.130\,(+2.95)}$ & $\mathbf{.128\text{--}.141\,(+2.34\text{--}+6.30)}$ & $.124\,(+1.17)$ & $\mathbf{.149\,(+8.64)}$ & $.123\,(+0.92)$ \\
Corruption-derived & $1000$ & $.356\pm.0590$ & $\mathbf{.401\,(+0.77)}$ & $\mathbf{.391\text{--}.408\,(+0.59\text{--}+0.88)}$ & $.253\,(-1.75)$ & $\mathbf{.412\,(+0.94)}$ & $.273\,(-1.40)$ \\
Corruption-derived & $1280$ & $.456\pm.0118$ & $.465\,(+0.76)$ & $.456\text{--}.471\,(-0.03\text{--}+1.28)$ & $.472\,(+1.32)$ & $.471\,(+1.25)$ & $.449\,(-0.64)$ \\
\bottomrule
\end{tabular}}
\end{table}

\paragraph{Local-set recovery separates repair-derived indices from clean readout indices.}
Table~\ref{tab:local-set-recovery} shows that clean coordinate-token rankings transfer at $k=500$, fall below the high-variance random baseline at $k=1000$, and converge with other non-oracle sets near half-width. The parenthesized $z$ values tie each fixed, split-local, sample-local, sample-specific, or bottom-ranked row to the matched random baseline. Bottom-ranked controls are useful negative controls: they do not explain the strong $k=500$ sample-specific corruption result or the $k=1000$ corruption-derived recovery. The corruption-derived block is cleaner for hard object masking: the fixed population ranking is above random at $k=500$, nearly matches the sample-specific corruption-derived row at $k=1000$, and then enters the same capacity-dominated regime.

\begin{table}[t]
\centering
\caption{Partial-occlusion local-set recovery with explicit index-set terminology. A rectangular mean-color patch covers $40\%$ of the ground-truth box. Entries report mean $R_{50}$ and mean within-split $z$ against ten random controls; Panel~A also reports Jaccard overlap $J$ with the fixed global coordinate-token Top-$k$ set. A fixed global or fixed population set is reused for every test sample; only the score used to construct that fixed set differs by panel.}
\label{tab:partial-occlusion}
\scriptsize
\setlength{\tabcolsep}{2.5pt}
\resizebox{\textwidth}{!}{%
\begin{tabular}{lrrrrrr}
\toprule
\multicolumn{7}{c}{Panel A: coordinate-token ranking family and matched alternatives under partial occlusion} \\
\midrule
$k$ & Random-$k$ & Fixed global & Split-local mean & Sample-local readout & Sample-specific corruption & Bottom-ranked \\
\midrule
$500$  & $.5101$ & $.5072\,(-1.00;\,J=1.000)$ & $.5088\,(-.48;\,J=.968)$ & $\mathbf{.5140\,(+1.32;\,J=.072)}$ & $.5073\,(-1.05;\,J=.257)$ & $.5078\,(-.78;\,J=.000)$ \\
$1000$ & $.5208$ & $\mathbf{.5032\,(-2.53;\,J=1.000)}$ & $\mathbf{.5022\,(-2.64;\,J=.982)}$ & $.5232\,(+.37;\,J=.223)$ & $\mathbf{.5108\,(-1.41;\,J=.398)}$ & $.5241\,(+.50;\,J=.000)$ \\
$1280$ & $.5212$ & $\mathbf{.5289\,(+2.42;\,J=1.000)}$ & $\mathbf{.5279\,(+2.00;\,J=.982)}$ & $\mathbf{.5313\,(+3.24;\,J=.314)}$ & $.5202\,(-.43;\,J=.470)$ & $.5227\,(+.85;\,J=.000)$ \\
\bottomrule
\end{tabular}}
\vspace{1mm}
\resizebox{\textwidth}{!}{%
\begin{tabular}{lrrrrr}
\toprule
\multicolumn{6}{c}{Panel B: population corruption-derived ranking family and matched alternatives under partial occlusion} \\
\midrule
$k$ & Random-$k$ & Fixed population & Sample-local readout & Sample-specific corruption & Bottom-ranked \\
\midrule
$500$  & $.5008$ & $.4972\,(-1.30\pm1.57)$ & $\mathbf{.5075\,(+2.12\pm.97)}$ & $.4994\,(-.57\pm.87)$ & $\mathbf{.5038\,(+1.06\pm1.01)}$ \\
$1000$ & $.5169$ & $\mathbf{.5059\,(-1.47\pm.88)}$ & $\mathbf{.5259\,(+1.37\pm1.14)}$ & $\mathbf{.5061\,(-1.52\pm.70)}$ & $.5206\,(+.51\pm.39)$ \\
$1280$ & $.5179$ & $.5178\,(+.02\pm.76)$ & $\mathbf{.5338\,(+3.77\pm1.48)}$ & $.5196\,(+.50\pm.83)$ & $.5232\,(+1.20\pm.52)$ \\
\bottomrule
\end{tabular}}
\begin{flushleft}
\footnotesize\textit{Set definitions.} Panel~A uses coordinate-token readout rankings: the fixed global set is one Top-$k$ set reused for every test sample, the split-local set is fixed per split, the sample-local readout set is recomputed from each clean test-sample coordinate-token trace, the sample-specific corruption set is recomputed from that sample's partial-occlusion clean--corrupt disruption, and the bottom-ranked set is the Bottom-$k$ set from the fixed global coordinate-token ranking. Panel~B uses population corruption-derived rankings: the fixed population set is the Top-$k$ set from the population average of per-sample disruption scores; the sample-specific corruption set uses the same underlying score but recomputes it for each sample from $F_{\theta}(h^{\mathrm{clean}})-h^{\mathrm{corrupt}}$ (here $\theta=-\infty$, so $F_{\theta}(h^{\mathrm{clean}})=h^{\mathrm{clean}}$); and the bottom-ranked set is the Bottom-$k$ set from the same population ranking.
\end{flushleft}
\end{table}

\paragraph{Partial occlusion separates index-set overlap from recovery.}
Table~\ref{tab:partial-occlusion} shows budget-dependent recovery rather than a simple overlap story. In Panel~A, fixed global and split-local coordinate-token sets have near-perfect overlap with the global set, yet both are below random at $k=1000$; the sample-specific corruption-derived row is also sub-random. This deficit is not merely parseability: for the fixed global row at $k=1000$, parseability is lower than random ($92.7\%$ versus $95.2\%$), but conditional localization is also lower ($R_{50}\mid\mathrm{parse}=.5425$ versus $.5470$). At $k=1280$, fixed global, split-local, and sample-local readout rows turn positive, while the sample-specific corruption-derived row is effectively neutral. The $J$ entries explain why this is not a rank-overlap story: high-overlap fixed/split sets can be sub-random, while low-overlap sample-local readout can be positive.

Panel~B contrasts a fixed population-average disruption ranking with the corresponding sample-specific disruption ranking. Both fail to beat random at $k=500$, both are sub-random at $k=1000$, and both return to near-null at $k=1280$. Across both panels, sample-local readout is the most consistently positive condition, with the strongest effect at half-width. Bottom-ranked controls are weakly positive for the population corruption-derived family, suggesting that avoiding highly disrupted dimensions can be safer than restoring them under partial occlusion. Supp. Table~\ref{tab:partial-severity} compares occlusion fractions at $k=1000$ and $k=1280$ and shows the same regime change directly: coordinate-token fixed and split-local sets are negative at $k=1000$ but nonnegative at $k=1280$, while random restoration flips from positive at $20\%$--$40\%$ occlusion to nonpositive at half-width.

\paragraph{Edge-attribution patching gives a path-level precision/recall check.}
Supp. Table~\ref{tab:eap-ablation} ablates the top scored attention and MLP edges under the $40\%$ partial-occlusion setting. Up to $100$ ablated edges, coordinate-token Top-$k$ edges are slightly more behaviorally concentrated than the complement and random controls ($\Delta R_{50}=-.034$ versus $-.029$ and $-.032$), supporting high precision for the coordinate-token endpoint objective. At $200$ edges the ordering reverses: complement and random edge sets damage detection more ($-.192$ and $-.166$) than the coordinate-token Top-$k$ set ($-.097$), even though the coordinate-token Top-$k$ set produces the largest change in $J_{\Delta}$. Supp. Fig.~\ref{fig:eap-heatmaps} is consistent with this interpretation: coordinate-token Top-$k$ paths concentrate more on output-proximal MLP structure, while broader detection-critical partial-occlusion behavior also uses complement and random-controlled pathways. Thus, edge attribution reinforces the dimension-level conclusion: endpoint readout is a useful circuit prior, but not a high-recall recovery explanation.

Supplementary Table~\ref{tab:parseability} shows that the main effects are not merely parseability artifacts. Qwen layer-34 Top flooring lowers both parseability and conditional localization, while Kimi-VL layer-26 and final-site effects mostly reflect formatting because $R_{50}\mid\mathrm{parse}$ stays near $.62$ across interventions. Object-mask endpoint recovery shows the strongest format collapse: the corrupted baseline is only $62\%$ parseable, but any $k=500$ restoration raises parseability to about $97\%$, indicating a global activation-repair effect rather than ranked-set-specific formatting. Partial occlusion shows only mild formatting degradation, and the $k=1000$ coordinate-token deficit persists after conditioning on parseable outputs.

\begin{table}[t]
\centering
\caption{Quasi-layer RMSNorm control using \texttt{bbox\_2d} readout rankings. The table compares repair rankings before/after making RMSNorm explicit, and endpoint-readout rankings against repair rankings in post-norm and transported pre-norm spaces.}
\label{tab:rmsnorm-main}
\scriptsize
\setlength{\tabcolsep}{4pt}
\resizebox{\textwidth}{!}{%
\begin{tabular}{lrrrr}
\toprule
$k$ & Repair pre/post overlap & Post endpoint-repair & Pre endpoint-repair & Pre-post \\
\midrule
$500$  & $\mathbf{.942\pm.008}$ & $.705\pm.013$ & $.713\pm.015$ & $\mathbf{+.008}$ \\
$1000$ & $\mathbf{.961\pm.004}$ & $.843\pm.008$ & $.843\pm.009$ & $\mathbf{+.001}$ \\
$1280$ & $\mathbf{.970\pm.003}$ & $.882\pm.006$ & $.877\pm.004$ & $\mathbf{-.004}$ \\
\bottomrule
\end{tabular}}
\end{table}

\paragraph{The quasi-layer control does not close the endpoint-repair gap.}
Table~\ref{tab:rmsnorm-main} tests whether the final RMSNorm boundary explains the readout/recovery gap. Repair rankings are almost unchanged across the boundary: pre/post repair overlap is $.942\pm.008$ at $k=500$ and $.970\pm.003$ at $k=1280$. Transporting endpoint attribution through the RMSNorm Jacobian changes endpoint-repair overlap by only $+.008$, $+.001$, and $-.004$ at $k=500,1000,1280$. The dissociation therefore remains visible in both post-norm and pre-norm spaces.

\section{Limitations and Future Work}
\label{sec:limitations}

This study is endpoint-centered. The interventions test whether selected residual-stream dimensions are load-bearing for coordinate emission or can transfer clean localization behavior into a corrupted decode trace; they do not identify the full upstream visual pathway. Vision-encoder representations, cross-modal attention, and intermediate MLP computations may contain repair-relevant structure. The depth grid is targeted rather than exhaustive: Tables~\ref{tab:coord-depth} and~\ref{tab:corruption-derived-depth} cover ten Qwen sites spanning early, mid, late, and final regimes, while Table~\ref{tab:kimivl-depth} covers proportional Kimi sites. Layer 8 in Table~\ref{tab:coord-depth} shows that $z$ values must be read against the random-control baseline, not raw $\Delta R_{50}$ alone. Full-flooring drops at layers 16, 17, 21, 24, and 32 show that fixed $\theta=-0.5$, calibrated as a non-collapsing final-site operating point, can be destructive at mid-network sites. We therefore separate full-flooring collapse, random-control shifts, and Top-$k$ selectivity rather than treating every depth effect as a discovered circuit.

The corruption settings are probes, not complete robustness benchmarks. Object masking removes local evidence, but is harder and less naturalistic than many real occlusions. Partial occlusion is closer to deployment because visible object evidence remains, but it is still synthetic: each occluded image is generated once per sample and held fixed across all $k$, mode, and split conditions. Conclusions should therefore be read as mechanistic evidence under occlusion-like damage, not as deployment claims about natural occluders, clutter, motion, or multi-instance scenes.

Future work can turn endpoint recovery into an annotation-free test-time repair module. One route is visible-region-conditioned pseudo-clean completion: an occlusion trigger detects likely partial visibility from box instability, low localization confidence, partial-object cues, or hidden-state anomaly scores; a completion stage generates candidates from visible object regions using symmetry, patch propagation, retrieval, inpainting, or feature-space completion; and a stability filter patches or reweights only recovery-relevant dimensions. The results suggest this module should be gated and conservative. Coordinate-token rankings are useful circuit priors, especially when some object evidence remains, but not universal repair sets. Future work should test whether pseudo-clean activation sourcing plus occlusion triggers and stability filtering across completions improves recovery without ground-truth boxes at deployment time.

\section{Discussion and Conclusion}
\label{sec:conclusion}

Endpoint readout is a useful starting point for VLM grounding analysis, but it is not a repair explanation by itself. The \texttt{bbox\_2d} field gives a stable, parseable localization endpoint, and coordinate-token triage targets the emitted coordinate values more directly. Corruption-derived rankings answer a different question: which dimensions repeatedly support recovery when object evidence is damaged.

The results make this separation concrete. Coordinate-token rankings are load-bearing for clean coordinate generation only after the layer-24-to-layer-32 transition in Qwen3-VL, with peak specificity at layer 34, and the Kimi-VL diagnostic shows a matching output-proximal transition in a second VLM, with onset at layer 24 and full selectivity by layer 26/final. Population corruption-derived rankings follow a different trajectory, with harmful mid-layer selectivity across layers 16--24 and a late beneficial regime at layer 35/final. Yet hard object-mask recovery is not explained by either clean readout geometry alone: half-width restoration is largely capacity-driven, while the strongest lower-budget repair evidence comes from corruption-derived indices. Partial occlusion adds the caution that high-overlap readout-derived dimension lists can still fail as recovery interventions: coordinate-token fixed and split-local sets fall below random at $k=1000$ but become beneficial at half-width, while population corruption-derived sets remain null or sub-random across the tested budgets. The edge-attribution patching diagnostic strengthens the same caution at the path level: readout-targeted edges are precise at small budgets but do not cover the broader set of detection-critical pathways. The RMSNorm control further narrows the explanation, since making the final normalization boundary explicit does not make endpoint readout substantially more repair-like.

The central contribution is therefore methodological. Endpoint rankings are principled circuit priors for localization, not sufficient evidence that the visual grounding pathway has been identified. Recovery from damaged visual evidence requires corruption-specific causal tests rather than endpoint-readout evidence alone.

\bibliographystyle{splncs04}

\clearpage
\appendix

\section{Supplementary Methods}
\label{app:method-details}

\subsection{Notation summary}
\label{app:notation-summary}

We use $i$ for sample index, $t$ for autoregressive decode-token position, $q$ for vocabulary-token index, $j$ and $p$ for hidden-dimension indices, and $\ell$ for decoder-layer index. The hidden width is $d=2560$. A ranking of hidden dimensions is denoted by $r=(r_1,\ldots,r_d)$. A selected intervention set of hidden dimensions is denoted by $I\subseteq\{1,\ldots,d\}$. For projector-based visualization, we use $S\subseteq\{1,\ldots,d\}$ to denote a selected dimension set. Intervention conditions are denoted by $c$.

\paragraph{Supplementary code.}
The anonymized supplementary material includes code for reproducing the intervention pipelines, table aggregation, and figure generation.

\subsection{Parseability convention}
\label{app:parseability}

Because the model expresses localization through generated text, we separate output-format validity from geometric localization accuracy. A response is considered parseable if it contains an identifiable \texttt{bbox\_2d} field followed by a coordinate list that can be converted into one or more four-coordinate boxes. Coordinates must be numeric after token detokenization and must define a valid box after the same post-processing used for all conditions, including coordinate conversion, coordinate ordering, and clipping to image bounds where applicable.

Parseability is imposed at two points. First, examples used to construct clean readout rankings are required to have a parseable clean localization response, because the ranking procedure needs localization-related target positions, such as the \texttt{bbox\_2d} field tokens or the generated coordinate tokens. Second, during intervention evaluation, parseability is not imposed by constrained decoding and outputs are not filtered post hoc. If an intervention causes the model to omit the \texttt{bbox\_2d} field, produce malformed coordinates, or generate a coordinate list that cannot be converted into a valid box, that trial is counted as unparseable. We report
\begin{equation}
\mathrm{Parseability}
=
\frac{1}{|\mathcal{D}_{\mathrm{eval}}|}
\sum_{i\in\mathcal{D}_{\mathrm{eval}}}
\mathbbm{1}\{\mathrm{parseable}(i)\}.
\end{equation}
For $R_{50}$ evaluation, unparseable outputs receive no successful match and therefore contribute zero localization recall. This convention prevents format collapse from being mistaken for a spatial error while also ensuring that endpoint recovery methods are not credited unless they recover a usable localization output.

Table~\ref{tab:parseability} applies this convention to representative regimes by reporting overall $R_{50}$, parseability, and $R_{50}\mid\mathrm{parse}$. The table covers clean-input depth triage, cross-model Kimi-VL depth triage, object-mask endpoint recovery, and partial-occlusion local-set recovery, separating output-format failure from spatial localization error.

\begin{table}[t]
\centering
\caption{Parseability decomposition for representative intervention regimes. $R_{50}$ counts unparseable outputs as zero; $R_{50}\mid\mathrm{parse}$ evaluates only parseable generations. Kimi-VL-A3B uses a bracketed normalized-coordinate format rather than Qwen's \texttt{bbox\_2d} JSON convention, so its baseline parseable rate is lower than Qwen's in these diagnostics.}
\label{tab:parseability}
\scriptsize
\setlength{\tabcolsep}{3pt}
\resizebox{\textwidth}{!}{%
\begin{tabular}{llcccc}
\toprule
\textbf{Experiment} & \textbf{Condition} & $R_{50}$ & Parse\,\% & $R_{50}\mid\mathrm{parse}$ & \textit{Evaluation cells} \\
\midrule
\multicolumn{6}{l}{\textit{Qwen coordinate-token depth triage, layer 34 ($k=1280$, $n\approx750$ per split)}} \\[1pt]
 & Baseline (no flooring)        & $.568$ & $97.9$ & $.580$ & $3$ held-out splits \\
 & Top coordinate-token floored  & $\mathbf{.408}$ & $\mathbf{89.5}$ & $\mathbf{.456}$ & $3$ held-out splits \\
 & Complement floored            & $.558$ & $97.0$ & $.575$ & $3$ held-out splits \\
 & Random, size-matched          & $.568$ & $97.9$ & $.580$ & $3$ held-out splits \\
\midrule
\multicolumn{6}{l}{\textit{Kimi-VL-A3B coordinate-token depth triage ($k=1024$, $n=735$--$767$)}} \\[1pt]
 & Baseline (no flooring)        & $.211$ & $33.5$ & $.628$ & $2$ held-out splits \\
 & Layer 26 Top                  & $\mathbf{.186}$ & $\mathbf{29.9}$ & $.622$ & $1$ held-out split \\
 & Layer 26 random               & $.226$ & $36.1$ & $.626$ & $1$ held-out split \\
 & Final Top                     & $\mathbf{.179}$ & $\mathbf{29.4}$ & $.610$ & $2$ held-out splits \\
 & Final random                  & $.223$ & $35.6$ & $.627$ & $2$ held-out splits \\
\midrule
\multicolumn{6}{l}{\textit{Object-mask endpoint recovery ($k=500$, $n\approx767$ per split)}} \\[1pt]
 & Corrupted baseline (no restoration)      & $\mathbf{.074}$ & $\mathbf{62.0}$ & $\mathbf{.120}$ & $5$ held-out splits \\
 & Coordinate-token $k=500$ restored        & $.540$ & $97.0$ & $.557$ & $5$ held-out splits \\
 & Population corruption-derived $k=500$ restored & $.530$ & $97.3$ & $.545$ & $5$ held-out splits \\
 & Random $k=500$ restored                  & $.542$ & $97.1$ & $.558$ & $5$ held-out splits \\
\midrule
\multicolumn{6}{l}{\textit{Partial occlusion, local index-set recovery ($k=1000$, $n=720$ per split)}} \\[1pt]
 & Fixed global coordinate-token & $\mathbf{.503}$ & $\mathbf{92.7}$ & $\mathbf{.5425}$ & $5$ held-out splits \\
 & Sample-local readout          & $\mathbf{.523}$ & $94.5$ & $\mathbf{.5538}$ & $5$ held-out splits \\
 & Random, size-matched          & $.521$ & $95.2$ & $.5470$ & $5$ held-out splits \\
\bottomrule
\end{tabular}}
\end{table}

Cells with one held-out split are treated as effect-size diagnostics. Table~\ref{tab:parseability} is intended to separate output-format failure from spatial localization degradation among parseable generations, not to replace the primary $R_{50}$ metric used in the main results. The object-mask recovery rows show that restoring any $k=500$ endpoint set largely repairs parseability, whereas the partial-occlusion rows show that localization error dominates once the object remains partly visible.

\subsection{Sites, token positions, and hook convention}
\label{app:sites-targets}

The final readout site is the output of the final RMSNorm immediately before the LM head. Decoder-layer sites are the outputs of the corresponding decoder blocks. For decoder layer $\ell$, the hook is attached to the output of \texttt{layers[$\ell$]}; for the final site, it is attached after final RMSNorm and before the LM head.

Hooks are decode-only. They fire during autoregressive generation steps with sequence length one and do not alter the image/prompt prefill pass. Target positions are generated tokens associated with localization: the token sequence spelling \texttt{bbox\_2d}, the coordinate-value tokens that follow it, or both, depending on the ranking being constructed. Thus, $t$ always indexes generated decode positions, not image-token positions.

\subsection{Kimi-VL coordinate-token depth-triage diagnostic}
\label{app:kimivl-coordinate-triage}

Kimi-VL-A3B does not use Qwen3-VL's \texttt{bbox\_2d} JSON convention in the evaluated generations. We prompt the model to localize the named object and return bounding boxes inside brackets; the evaluated generations use plain normalized boxes of the form \texttt{[x1, y1, x2, y2]}. The Kimi diagnostic therefore uses coordinate-token triage directly. At each bounding-box decode step, we construct a token-specific contrast direction from the untied language-model head,
\begin{equation}
w_t = W_{\mathrm{lm}}[:,y_t] - \frac{1}{|\mathcal{N}_t|}\sum_{n\in\mathcal{N}_t} W_{\mathrm{lm}}[:,n],
\end{equation}
and score each hidden dimension by
\begin{equation}
s_{t,j}^{\mathrm{Kimi}} = \left|F_{\theta}(h_{t,j})-h_{t,j}\right|\, |w_{t,j}|,
\qquad \theta=-0.5.
\end{equation}
Scores are aggregated hierarchically from decode step to coordinate value, box, sample, and global split ranking. The resulting coordinate-value ranking has 2048 dimensions and supplies the Top-$k$ input for Kimi-VL depth triage.

The reported Kimi-VL experiment uses $k=1024$, the half-width counterpart of Qwen3-VL's $k=1280$, and evaluates proportional depth sites in Kimi's 27-layer decoder, including layers 6, 12, 13, 16, 18, 24, 25, 26, and the final norm output. The intervention is applied at decoder-block outputs for layer sites and at the final norm output for the final site. The Kimi table in the main text reports coordinate-token depth triage only. A Kimi population corruption-derived ranking would require a corrected, variance-normalized repair score because the naive donor-minus-corrupt displacement is dominated by outlier channels; this recovery-side extension is outside the reported Kimi diagnostic.

\subsection{Contrast readout directions}
\label{app:readout-directions}

Let $h_{i,t}\in\mathbb{R}^{d}$ be the hidden state for sample $i$ at localization-related decode position $t$. Let $W_q\in\mathbb{R}^{d}$ denote the LM-head, or unembedding, vector associated with vocabulary token $q$. We orient $W_q$ so that the token-$q$ logit contribution from hidden state $h_{i,t}$ is $h_{i,t}^{\top}W_q$ up to any token-specific bias.

For a target token $y_{i,t}$, let
\begin{equation}
\mathcal{N}_{i,t}
=
\{n_{i,t,1},\ldots,n_{i,t,M_{i,t}}\}
\end{equation}
be a set of $M_{i,t}$ negative alternative tokens appropriate to the same output position. The index $m$ below is only a counter over these negative alternatives. The average negative-token direction is
\begin{equation}
\overline W^{-}_{i,t}
=
\frac{1}{M_{i,t}}
\sum_{m=1}^{M_{i,t}}
W_{n_{i,t,m}} .
\end{equation}
The contrast readout direction is then
\begin{equation}
w_{i,t}=W_{y_{i,t}}-\overline W^{-}_{i,t}.
\end{equation}
Thus, $w_{i,t}$ points toward the target token relative to the average competing-token direction. The scalar $w_{i,t,j}$ is the contrastive logit-facing direction for hidden dimension $j$. For \texttt{bbox\_2d} rankings, $y_{i,t}$ is drawn from the stable field-token positions. For coordinate-token rankings, $y_{i,t}$ is the sample-specific coordinate token generated for the particular image and object.

\begin{table}[t]
\centering
\caption{Negative-set sensitivity of coordinate-token rankings. Entries report Jaccard overlap between Top-$k$ hidden-dimension sets produced by the main position-specific negative pool and rankings produced by alternative negative-token pools. The check uses one held-out split with $n=147$ valid generations. The first row compares a small-sample replica of the main-pool ranking with the corresponding full-scale main-pool ranking, isolating negative-pool effects from sample-size noise.}
\label{tab:negpool-sensitivity}
\small
\setlength{\tabcolsep}{5pt}
\begin{tabular}{lccc}
\toprule
Pool comparison & $k=500$ & $k=1000$ & $k=1280$ \\
\midrule
main pool replica vs. full ranking & $\mathbf{.953}$ & $\mathbf{.970}$ & $\mathbf{.966}$ \\
\midrule
main pool vs. digit-only pool & $\mathbf{.244}$ & $\mathbf{.410}$ & $\mathbf{.502}$ \\
main pool vs. coordinate-format pool & $\mathbf{.439}$ & $\mathbf{.548}$ & $\mathbf{.614}$ \\
digit-only pool vs. coordinate-format pool & $\mathbf{.247}$ & $\mathbf{.428}$ & $\mathbf{.492}$ \\
\bottomrule
\end{tabular}
\end{table}

Table~\ref{tab:negpool-sensitivity} directly tests sensitivity to the negative-token pool in the contrastive readout direction. The main-pool replica closely matches the full-scale main-pool ranking (Jaccard at least $.95$ at every reported $k$), so the alternative-pool differences are not explained by the smaller sensitivity-check sample. The alternative pools yield moderate rather than near-total Top-$k$ overlap ($.24$--$.61$, increasing with $k$). Thus, coordinate-token dimension identities are partly negative-pool-dependent. We therefore treat the negative pool as part of the ranking specification and rely on held-out intervention behavior in the main depth-triage and recovery tables, rather than exact rank identity alone, for the paper's causal claims.

\subsection{Corruption-derived ranking}
\label{app:corruption-derived-ranking}

The corruption-derived ranking is built from object-mask repair displacements rather than clean-only readout contributions. For sample $i$, the corruption-derived score for hidden dimension $j$ is
\begin{equation}
s^{\mathrm{der}}_{i,j}
=
\frac{1}{T_i}
\sum_{t=1}^{T_i}
\left|
\left(
h^{\mathrm{clean}}_{i,t,j}
-
h^{\mathrm{mask}}_{i,t,j}
\right)
 w_{i,t,j}
\right|,
\end{equation}
where $h^{\mathrm{mask}}$ is the trace under object-mask corruption. Each sample produces a full $d$-dimensional corruption-derived ordering. Unlike the readout score, $s^{\mathrm{der}}_{i,j}$ does not involve the flooring operator $F_{\theta}$; it uses the raw clean-to-object-mask endpoint displacement projected onto the readout direction, so $\theta$ is not calibrated or used for this ranking.

When raw scores are available, we aggregate them by mean score over the ranking population. When only rank order is retained, we aggregate sample-level orderings by average rank position:
\begin{equation}
S^{\mathrm{der}}_{j}
=
\frac{1}{|\mathcal{D}_{\mathrm{rank}}|}
\sum_{i\in\mathcal{D}_{\mathrm{rank}}}
\operatorname{rank}_{i}(j).
\end{equation}
Dimensions persistently appearing near the top of sample-level corruption-derived rankings become high-ranked in the corruption-derived ranking. This distinguishes two kinds of global ranking: clean readout rankings capture output-interface leverage, whereas corruption-derived rankings capture recurring object-mask recovery leverage.

\subsection{Flooring motivation and readout-ranking score}
\label{app:flooring-motivation}

Elementwise flooring supplies the dimensionwise perturbation used to define readout scores and clean-input depth triage. For hidden coordinate $h_j$, the operator
\begin{equation}
F_{\theta}(h_j)=\max(h_j,\theta),\qquad \theta=-0.5
\end{equation}
induces the displacement
\begin{equation}
\Delta h_{i,t,j}
=
F_{\theta}(h^{\mathrm{clean}}_{i,t,j})
-
 h^{\mathrm{clean}}_{i,t,j}.
\end{equation}
Thus, dimensions already above the threshold are unchanged, whereas below-threshold coordinates are lifted to the calibrated floor. The threshold $\theta=-0.5$ is used as a non-collapsing operating point for Qwen3-VL-4B and Kimi-VL-A3B: it perturbs localization behavior while largely preserving parseable generation.

The readout-triage score projects this flooring displacement onto the contrast direction for the target localization token. For a clean readout ranking, the per-position contribution of hidden dimension $j$ is
\begin{equation}
s^{\mathrm{read}}_{i,t,j}
=
\left|
\Delta h_{i,t,j}\, w_{i,t,j}
\right| .
\end{equation}
Scores are averaged over samples and target positions in the ranking split:
\begin{equation}
S^{\mathrm{read}}_{j}
=
\frac{1}{|\mathcal{D}_{\mathrm{rank}}|}
\sum_{i\in\mathcal{D}_{\mathrm{rank}}}
\frac{1}{T_i}
\sum_{t=1}^{T_i}
 s^{\mathrm{read}}_{i,t,j}.
\end{equation}
Sorting dimensions by $S^{\mathrm{read}}_{j}$ yields a global final-site readout ranking. In the initial triage that motivated this study, the ranking was computed from a small diagnostic subset with balanced outcome types and then evaluated on larger image-cluster bootstrap splits by comparing Top-$k$, random-$k$, and complement-$k$ selective flooring conditions.

For that diagnostic check, let $R_{50}^{\mathrm{full\mbox{-}floor}}$ denote the score after flooring all dimensions at the final site, and let $R_{50}^{\mathrm{clean}}$ denote the unmodified clean baseline. For a selective flooring condition $c$, the fraction of the full final-site gain recovered is
\begin{equation}
\mathrm{GainFrac}^{\mathrm{floor}}(c,k)
=
100\cdot
\frac{
R_{50}^{\mathrm{floor}}(c,k)-R_{50}^{\mathrm{clean}}
}{
R_{50}^{\mathrm{full\mbox{-}floor}}-R_{50}^{\mathrm{clean}}
}.
\end{equation}
The observed pattern motivates the ranking as a final-readout triage prior: Top-$k$ dimensions recover more of the full final-site gain than size-matched random sets across rank sizes, while complement flooring removes or reverses the gain once the withheld Top-$k$ set is large enough. At very small $k$, complement sets can outperform Top-$k$ because the complement still contains most hidden dimensions; this is expected and does not by itself imply that the smallest Top-$k$ set is sufficient. These checks justify using the ranking as a stable endpoint reference, but they do not assume that endpoint-ranked dimensions are also sufficient for object-mask repair.

\subsection{Partial-occlusion corruption}
\label{app:partial-occlusion}

Partial occlusion is implemented as a softer visual corruption than object masking and is used as a synthetic occlusion-like probe for the practically common case in which the target remains partly visible. For each ground-truth box, the corruption samples a patch inside the box whose area equals a specified fraction of the box area. With occlusion fraction $\rho$, the square-side fraction is $\sqrt{\rho}$; for example, $\rho=.4$ covers approximately $40\%$ of the box area rather than $40\%$ of each side. The patch location is sampled uniformly subject to remaining inside the ground-truth box, so part of the target object remains visible. The patch is drawn as either a rectangle or ellipse and filled with a constant value, using the image mean in the reported experiments.

Partial occlusion is included as a corruption condition alongside full object masking, background corruption, and image shuffling. In $k$-scaling recovery and local-set recovery, each corrupted image is constructed once per sample and then held fixed. The same partial-occlusion image is reused across all rank sizes, index-set modes, random seeds, and split-transfer conditions for that sample. The corruption is therefore paired within an experiment: differences among modes reflect the activation restoration condition, not a different occlusion draw. The reported partial-occlusion sensitivity analyses use $\theta=-\infty$ with the same restoration operator, so floored restoration is algebraically equivalent to clean endpoint replacement. Exact replication across separate invocations requires holding the generated corruptions fixed or fixing the external image-generation seed.

\begin{table}[t]
\centering
\caption{Final partial-occlusion severity sweep comparing $k=1000$ and $k=1280$. Entries are $\Delta R_{50}$ offsets relative to the corresponding no-restoration partial-occlusion baseline over five held-out splits with $150$ samples per split. The ablation varies both restoration budget and occlusion area fraction while using the same generic index-set scopes as Table~\ref{tab:partial-occlusion}.}
\label{tab:partial-severity}
\scriptsize
\setlength{\tabcolsep}{2.6pt}
\resizebox{\textwidth}{!}{%
\begin{tabular}{lrrrrrr}
\toprule
Condition & \multicolumn{3}{c}{$k=1000$} & \multicolumn{3}{c}{$k=1280$} \\
\cmidrule(lr){2-4} \cmidrule(lr){5-7}
& $20\%$ & $40\%$ & $60\%$ & $20\%$ & $40\%$ & $60\%$ \\
\midrule
Fixed population corruption-derived & $-.004$ & $+.007$ & $+.077^{\dagger}$ & $+.002$ & $-.010^{\dagger}$ & $+.010$ \\
Fixed global coordinate-token & $-.032^{\dagger}$ & $-.043^{\dagger}$ & $-.097^{\dagger}$ & $+.010$ & $+.007$ & $+.012$ \\
Split-local population corruption-derived & $-.010$ & $+.011$ & $+.043^{\dagger}$ & $+.002$ & $-.009^{\ddagger}$ & $+.008$ \\
Split-local coordinate-token & $-.026^{\dagger}$ & $-.043^{\dagger}$ & $-.032^{\dagger}$ & $+.015^{\ddagger}$ & $+.007$ & $+.001$ \\
Sample-local readout & $+.016$ & $+.006$ & $+.088^{\dagger}$ & $+.002$ & $+.019$ & $+.013$ \\
Sample-specific corruption-derived & $-.013$ & $-.000$ & $-.049$ & $-.008$ & $-.013^{\dagger}$ & $+.002$ \\
Random-$k$ & $+.015^{\dagger}$ & $+.015^{\dagger}$ & $-.002$ & $-.007^{\dagger}$ & $-.002$ & $-.007$ \\
Bottom-ranked & $+.025^{\dagger}$ & $+.006$ & $-.044$ & $+.001$ & $+.022^{\ddagger}$ & $-.007$ \\
\bottomrule
\end{tabular}}
\begin{flushleft}
\footnotesize $\dagger$ marks a bootstrap $95\%$ confidence interval excluding zero; $\ddagger$ marks a very tight confidence interval excluding zero. The final sweep shows two coupled regime changes. First, coordinate-token fixed and split-local rows are consistently negative at $k=1000$ but nonnegative at $k=1280$. Second, random restoration is positive at $k=1000$ for $20\%$--$40\%$ occlusion but nonpositive at $k=1280$, indicating that the identity of restored dimensions matters once the restoration budget reaches half-width. At $40\%$ occlusion and $k=1280$, bottom-ranked dimensions are the strongest positive condition, while population corruption-derived and sample-specific corruption-derived sets are reliably harmful.
\end{flushleft}
\end{table}

\subsection{Dimension sets and intervention operators}
\label{app:dimension-sets}

Let $r=(r_1,\ldots,r_d)$ be a ranking of hidden dimensions. For rank size $k$, define
\begin{equation}
I_{\mathrm{top}}(k)=\{r_1,\ldots,r_k\},
\end{equation}
\begin{equation}
I_{\mathrm{bottom}}(k)=\{r_{d-k+1},\ldots,r_d\},
\end{equation}
\begin{equation}
I_{\mathrm{comp}}(k)=\{r_{k+1},\ldots,r_d\}.
\end{equation}
Random controls sample size-$k$ sets uniformly from $\{1,\ldots,d\}$, and complement-random controls sample size-$(d-k)$ sets. All sets are hidden-dimension sets, not token-position sets.

In clean selective flooring, selected dimensions are floored while all other dimensions are left unchanged:
\begin{equation}
h'_{t,I}=F_{\theta}(h_{t,I}),\qquad
h'_{t,\bar I}=h_{t,\bar I}.
\end{equation}
In endpoint recovery experiments, the image is first corrupted by masking the ground-truth object region with the per-image mean color. Selected corrupted dimensions are then replaced with their clean values:
\begin{equation}
h^{\mathrm{patch}}_{t,I}=h^{\mathrm{clean}}_{t,I},\qquad
h^{\mathrm{patch}}_{t,\bar I}=h^{\mathrm{mask}}_{t,\bar I}.
\end{equation}
This convention is important: in clean flooring experiments, a set label denotes the dimensions perturbed; in endpoint recovery experiments, a set label denotes the dimensions replaced by clean endpoint values.

\subsection{Layer-resolved selective flooring}
\label{app:layer-resolved-flooring}

To determine where ranked dimensions become load-bearing, we hold a ranking fixed and move the clean selective-flooring hook across decoder layers and the final site. For decoder layer $\ell$, the hook is attached to the output of \texttt{layers[$\ell$]}; for the final site, it is attached after final RMSNorm and before the LM head. We compare the flooring intervention conditions
\begin{equation}
\mathcal{C}_{\mathrm{floor}}
=
\{\mathrm{full},\mathrm{top}\text{-}k,\mathrm{comp},\mathrm{rand}\text{-}k,\mathrm{rand}\text{-}\mathrm{comp}\},
\end{equation}
where $\mathrm{top}\text{-}k$ floors the Top-$k$ dimensions, $\mathrm{comp}$ floors the complement of the Top-$k$ set, $\mathrm{rand}\text{-}k$ floors a size-$k$ random set, and $\mathrm{rand}\text{-}\mathrm{comp}$ floors a size-$(d-k)$ random set. We evaluate selected rank sizes, including $k=500$, $1000$, and $1280$.

For flooring condition $c\in\mathcal{C}_{\mathrm{floor}}$, decoder layer $\ell$, and rank size $k$, we report
\begin{equation}
\Delta R_{50}^{\mathrm{floor}}(c,\ell,k)
=
R_{50}^{\mathrm{floor}}(c,\ell,k)-R_{50}^{\mathrm{clean}}.
\end{equation}
We also compute the selectivity gap
\begin{equation}
\mathrm{gap}^{\mathrm{floor}}(\ell,k)
=
\Delta R_{50}^{\mathrm{floor}}(\mathrm{top}\text{-}k,\ell,k)
-
\Delta R_{50}^{\mathrm{floor}}(\mathrm{comp},\ell,k).
\end{equation}
A depth at which complement flooring becomes disruptive indicates that non-top dimensions are load-bearing at that site. Cross-$k$ comparisons localize whether the disruptive information lies inside the top ranks, outside them, or diffusely across the residual stream.

\subsection{$k$-scaling endpoint recovery}
\label{app:k-scaling}

To test whether visual-corruption endpoint recovery is ranking-enriched or capacity-driven, we replace selected dimension sets at the final site over a wide range of $k$ values, including the tested $k=500$ budgets, the half-width point $k=d/2=1280$, and near-full endpoint recovery. We compare the endpoint-recovery intervention conditions
\begin{equation}
\mathcal{C}_{\mathrm{rec}}
=
\{\mathrm{top}\text{-}k,\mathrm{bottom}\text{-}k,\mathrm{comp},\mathrm{rand}\text{-}k,\mathrm{rand}\text{-}\mathrm{comp},\mathrm{full}\}.
\end{equation}
Here, $\mathrm{top}\text{-}k$ replaces the Top-$k$ dimensions with their clean-trace values, $\mathrm{bottom}\text{-}k$ replaces the Bottom-$k$ dimensions, $\mathrm{comp}$ replaces the complement of the Top-$k$ set, $\mathrm{rand}\text{-}k$ replaces a size-$k$ random set, $\mathrm{rand}\text{-}\mathrm{comp}$ replaces a size-$(d-k)$ random set, and $\mathrm{full}$ replaces all endpoint dimensions with their clean-trace values.

For endpoint-recovery condition $c\in\mathcal{C}_{\mathrm{rec}}$, recovery is measured relative to the selected corruption baseline, object mask or partial occlusion, and the full-endpoint-recovery ceiling:
\begin{equation}
\Delta R_{50}^{\mathrm{rec}}(k,c)
=
R_{50}^{\mathrm{rec}}(k,c)-R_{50}^{\mathrm{mask}},
\end{equation}
\begin{equation}
\mathrm{Recovery}(k,c)
=
100\cdot
\frac{
R_{50}^{\mathrm{rec}}(k,c)-R_{50}^{\mathrm{mask}}
}{
R_{50}^{\mathrm{rec}}(k,\mathrm{full})-R_{50}^{\mathrm{mask}}
}.
\end{equation}
We summarize each curve by area under the $k$-curve, the smallest $k$ reaching half-maximal recovery, and slopes under log-capacity and linear-capacity fits. A positive Top-minus-random gap indicates ranking-enriched recovery. Top/random/bottom parity indicates that endpoint recovery depends mainly on how many endpoint dimensions are replaced rather than which endpoint-ranked dimensions are selected.

\subsection{Local-set endpoint recovery}
\label{app:local-transfer}

To test whether a global ranking transfers to individual corrupted examples, we compare five generic index-set scopes at $k\in\{500,1000,1280\}$: fixed global, split-local, sample-local readout, sample-specific corruption-derived, bottom-ranked, and random. A fixed global or fixed population set always means that the same Top-$k$ hidden dimensions are restored for every test sample. A split-local set is fixed within a split but may differ across split-specific rankings. A sample-local readout set or sample-specific corruption-derived set varies by test sample.

For coordinate-token local-set recovery, the fixed global set is the Top-$k$ set from the global coordinate-token readout ranking, computed once on a triage/ranking split and reused for all test samples. The split-local set uses a coordinate-token ranking recomputed on one split and then held fixed for the corresponding evaluation. The sample-local readout set recomputes the coordinate-token readout ranking from the current test sample's own clean readout trace. The sample-specific corruption-derived set selects the Top-$k$ dimensions most disrupted by that same sample's visual corruption, and the bottom-ranked set is the Bottom-$k$ set from the fixed global coordinate-token ranking.

For population corruption-derived local-set recovery, the fixed population set is the Top-$k$ set under the population average of per-sample disruption scores, so it is fixed for every held-out test sample. The sample-specific corruption-derived set uses the same underlying clean--corrupt disruption score but recomputes it from the current sample's traces, and therefore varies per sample. With the reported partial-occlusion setting, this score is computed from $F_{\theta}(h^{\mathrm{clean}})-h^{\mathrm{corrupt}}$; because $\theta=-\infty$, this reduces to $h^{\mathrm{clean}}-h^{\mathrm{corrupt}}$. The bottom-ranked set is the Bottom-$k$ set from the same population corruption-derived ranking. This experiment asks whether knowing which dimensions were disrupted on the current sample beats using the dimensions most often disrupted across the held-out population.

We compute Jaccard overlap between local Top-$k$ sets and global Top-$k$ sets to quantify rank heterogeneity. To avoid overloading $J$ with Jacobian or objective notation, we denote this set-overlap score by $\operatorname{Jac}$:
\begin{equation}
\operatorname{Jac}(I_i,I_{\mathrm{global}})
=
\frac{|I_i\cap I_{\mathrm{global}}|}
{|I_i\cup I_{\mathrm{global}}|}.
\end{equation}
Low overlap indicates that a fixed global clean-readout ranking may not capture the dimensions needed by individual corrupted examples.

\subsection{Edge-attribution patching under partial occlusion}
\label{app:eap-partial-occlusion}

This diagnostic complements endpoint dimension replacement by asking which attention and MLP edges route the coordinate-token endpoint clean--corrupt gap under partial occlusion. It is not a new recovery operator; it is a path-level validation of whether the coordinate-token endpoint objective isolates behaviorally important edges.

For a clean trace and a partial-occlusion trace, let $h^{\mathrm{clean}}_{t}$ and $h^{\mathrm{occ}}_{t}$ denote the final residual stream at localization-related decode position $t$. Let $S\subset\mathbb{R}^{2560}$ be a half-width endpoint subspace with $k=1280$ directions, and let $P_S$ be the corresponding coordinate projector. The endpoint-gap objective is
\begin{equation}
J_{\Delta}(S)
=
\sum_{t\in\mathcal{T}_{\mathrm{box}}}
\left\|P_S\left(h^{\mathrm{clean}}_{t}-h^{\mathrm{occ}}_{t}\right)\right\|_2^2 .
\end{equation}
We compare three controlled endpoint subspaces: the coordinate-token readout-ranked Top-$k$ dimensions, the complementary half of the residual stream, and a size-matched random half. The coordinate-token Top-$k$ and complement subspaces partition the full hidden width, while the random subspace is an orthogonal size-matched control. The three objectives share the same partial-occlusion forward pass for each sample before separate backward objectives score the same candidate edge pool.

For an edge activation $a_e$, the edge-attribution patching score is the first-order clean-to-corrupt contribution
\begin{equation}
\mathrm{EAP}(e;S)
\approx
\left(a^{\mathrm{clean}}_e-a^{\mathrm{occ}}_e\right)^{\top}
\nabla_{a_e} J_{\Delta}(S).
\end{equation}
The diagnostic evaluated $100$ candidate examples; $27$ met the validity criteria for edge-gradient scoring, yielding $37{,}800$ scored edges for each endpoint objective. The coordinate-token Top-$k$, complement, and random-$k$ objectives respectively produced $15{,}235$, $14{,}883$, and $14{,}860$ positive-score edges. For visualization, each score ranking is converted into a directed graph with scored edges plus bridge edges for connectivity. Boundary paths are extracted from visual and prompt source positions to the highest-scoring localization answer positions; the top paths are overlaid on layer-by-token heatmaps.

\begin{table}[t]
\centering
\caption{Edge-attribution patching validation under $40\%$ partial occlusion. The top-$n$ edges from each endpoint objective are zero-ablated at inference on $300$ held-out examples. $\Delta R_{50}$ is relative to the unablated partial-occlusion run, and $\Delta J_{\Delta}$ is the change in the corresponding endpoint-gap objective. Bold entries mark the most diagnostic cells discussed in the main text.}
\label{tab:eap-ablation}
\scriptsize
\setlength{\tabcolsep}{3pt}
\resizebox{\textwidth}{!}{%
\begin{tabular}{rrrrrrr}
\toprule
$n_{\mathrm{edges}}$ & $\Delta R_{50}$ Coordinate-token Top-$k$ & $\Delta R_{50}$ Complement & $\Delta R_{50}$ Random-$k$ & $\Delta J_{\Delta}$ Coordinate-token Top-$k$ & $\Delta J_{\Delta}$ Complement & $\Delta J_{\Delta}$ Random-$k$ \\
\midrule
$10$  & $-.002$ & $-.021$ & $-.010$ & $+3621$  & $+3208$  & $+3888$ \\
$25$  & $-.019$ & $-.018$ & $-.016$ & $+2321$  & $+2216$  & $+3618$ \\
$50$  & $-.020$ & $-.022$ & $-.019$ & $-5140$  & $-3763$  & $+2315$ \\
$100$ & $\mathbf{-.034}$ & $-.029$ & $-.032$ & $-10906$ & $-8799$  & $-13078$ \\
$200$ & $-.097$ & $\mathbf{-.192}$ & $\mathbf{-.166}$ & $\mathbf{-33548}$ & $-16025$ & $-15411$ \\
\bottomrule
\end{tabular}}
\end{table}

Table~\ref{tab:eap-ablation} shows two regimes. Up to $100$ ablated edges, the three edge rankings have similar behavioral impact, and the coordinate-token Top-$k$ objective gives the largest $R_{50}$ drop at $100$ edges. This is the high-precision regime: the strongest coordinate-token-targeted edges are also behaviorally concentrated. From $100$ to $200$ edges, the behavioral ordering reverses. The coordinate-token Top-$k$ edge set causes only an additional $-.063$ $R_{50}$ drop, while the complement and random edge sets add $-.163$ and $-.134$. At the same time, the coordinate-token Top-$k$ objective dominates the endpoint-gap change in absolute magnitude. Thus, the coordinate-token Top-$k$ EAP circuit is highly specific to $J_{\Delta}$ but has limited recall for the full detection circuit under partial occlusion.

\begin{figure}[t]
\centering
\includegraphics[width=.32\linewidth]{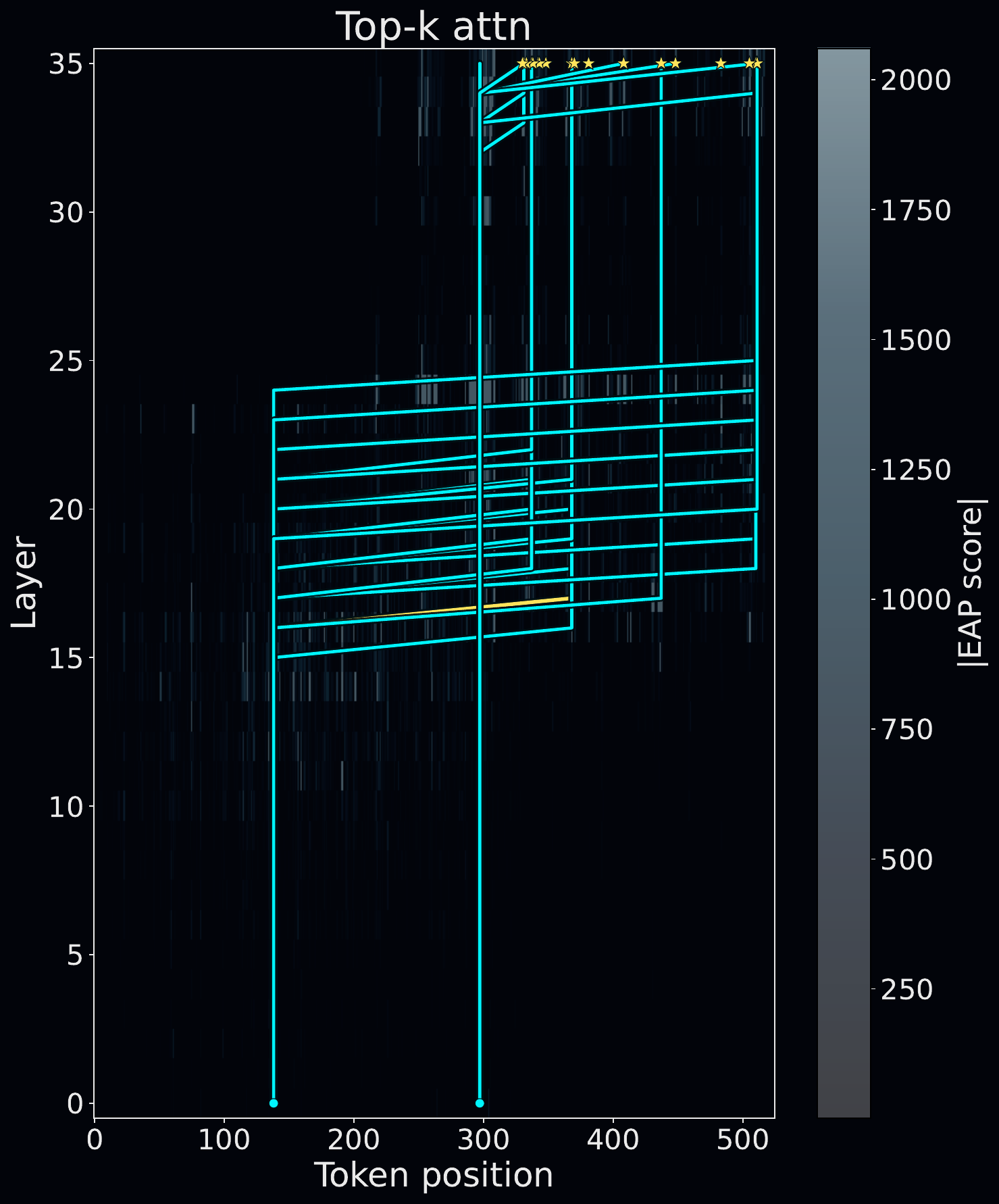}\hfill
\includegraphics[width=.32\linewidth]{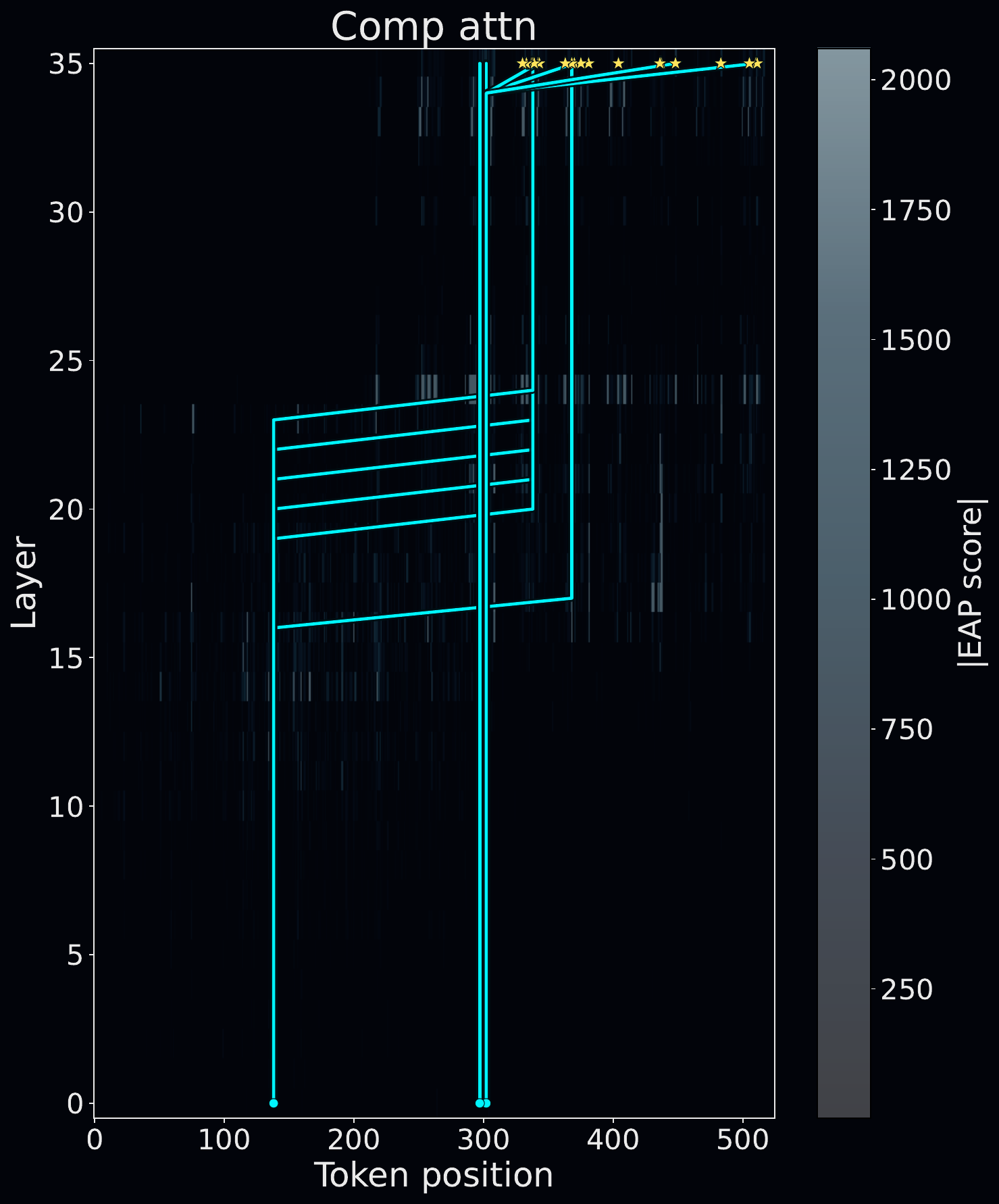}\hfill
\includegraphics[width=.32\linewidth]{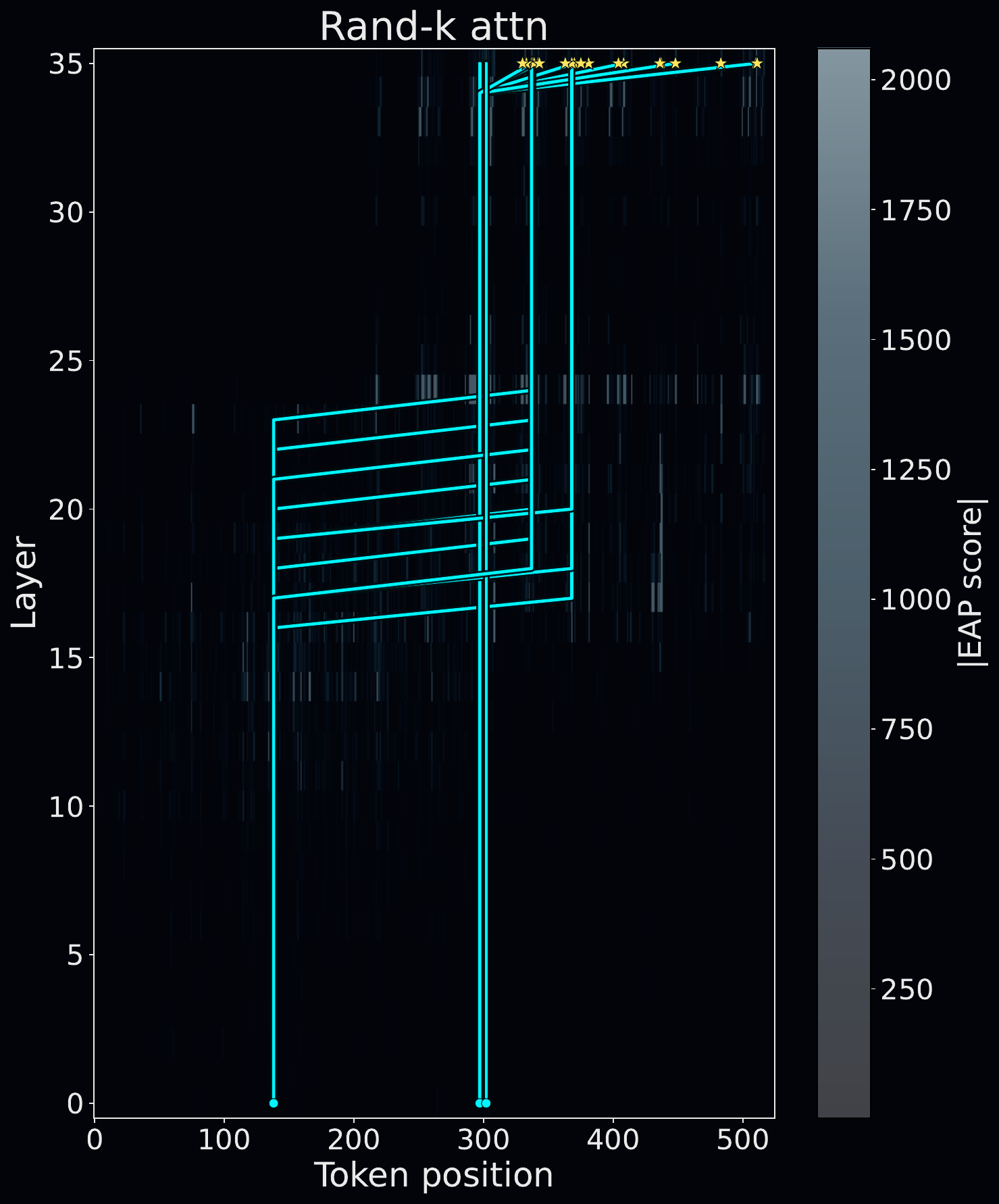}\\[1mm]
\includegraphics[width=.32\linewidth]{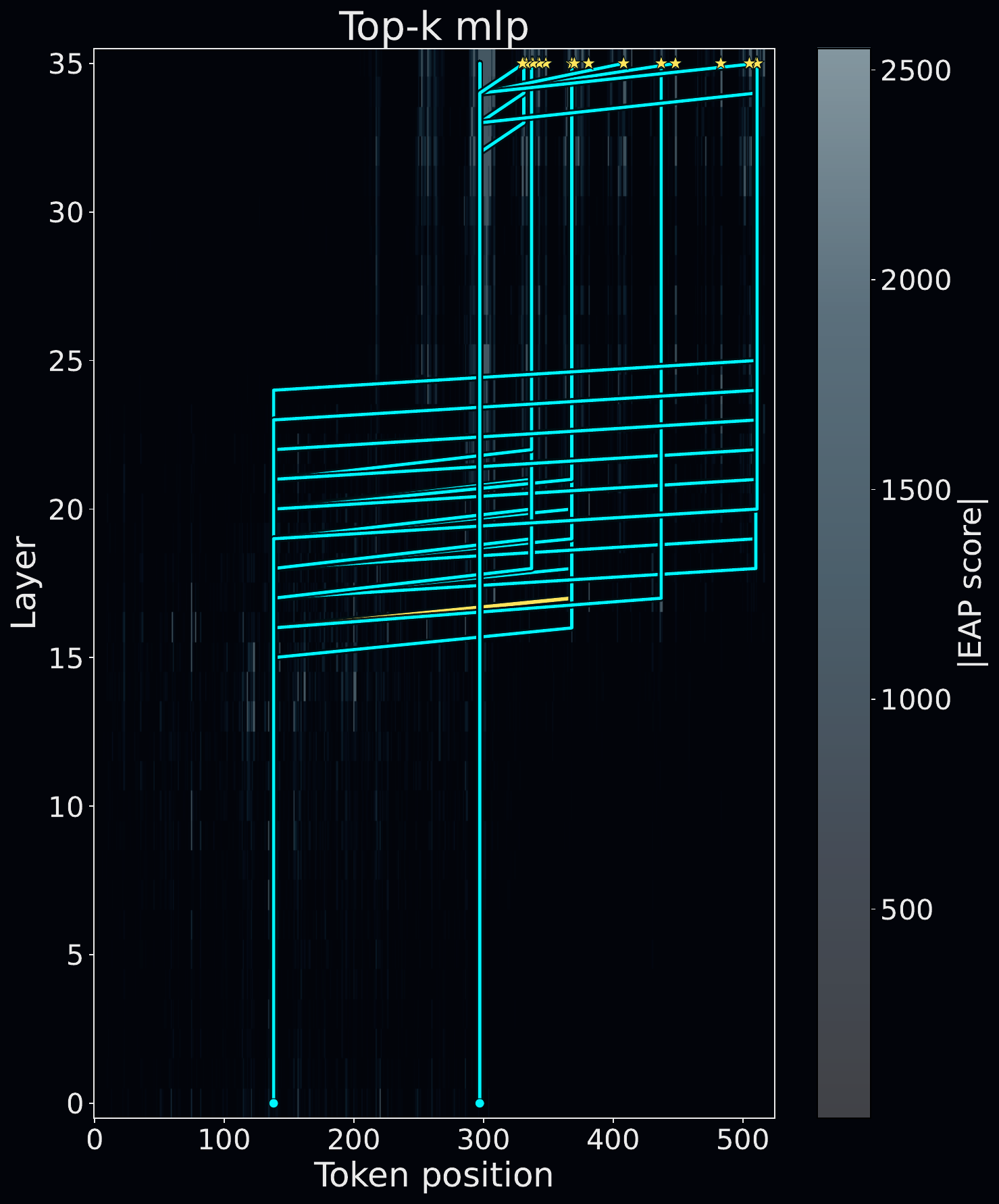}\hfill
\includegraphics[width=.32\linewidth]{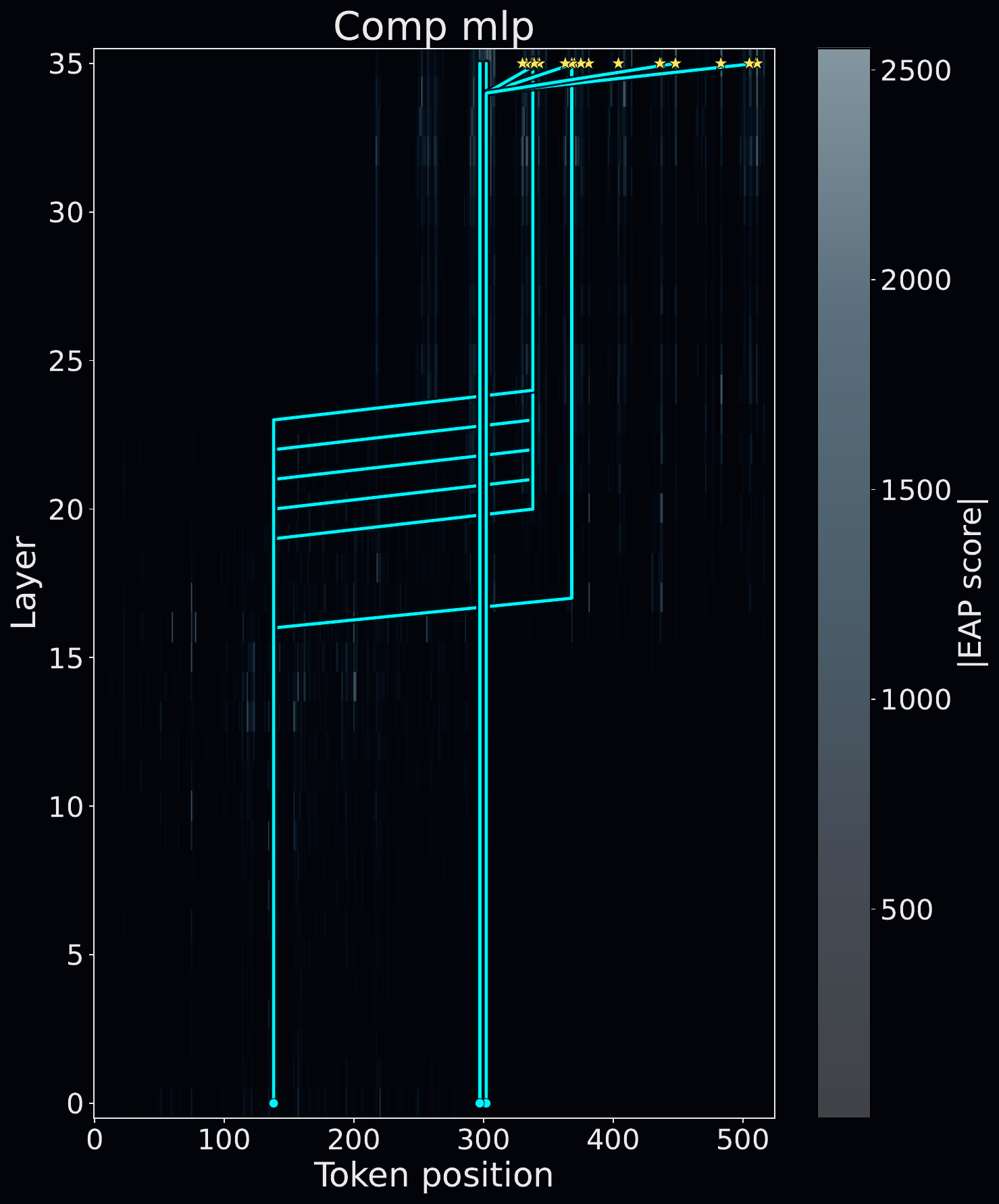}\hfill
\includegraphics[width=.32\linewidth]{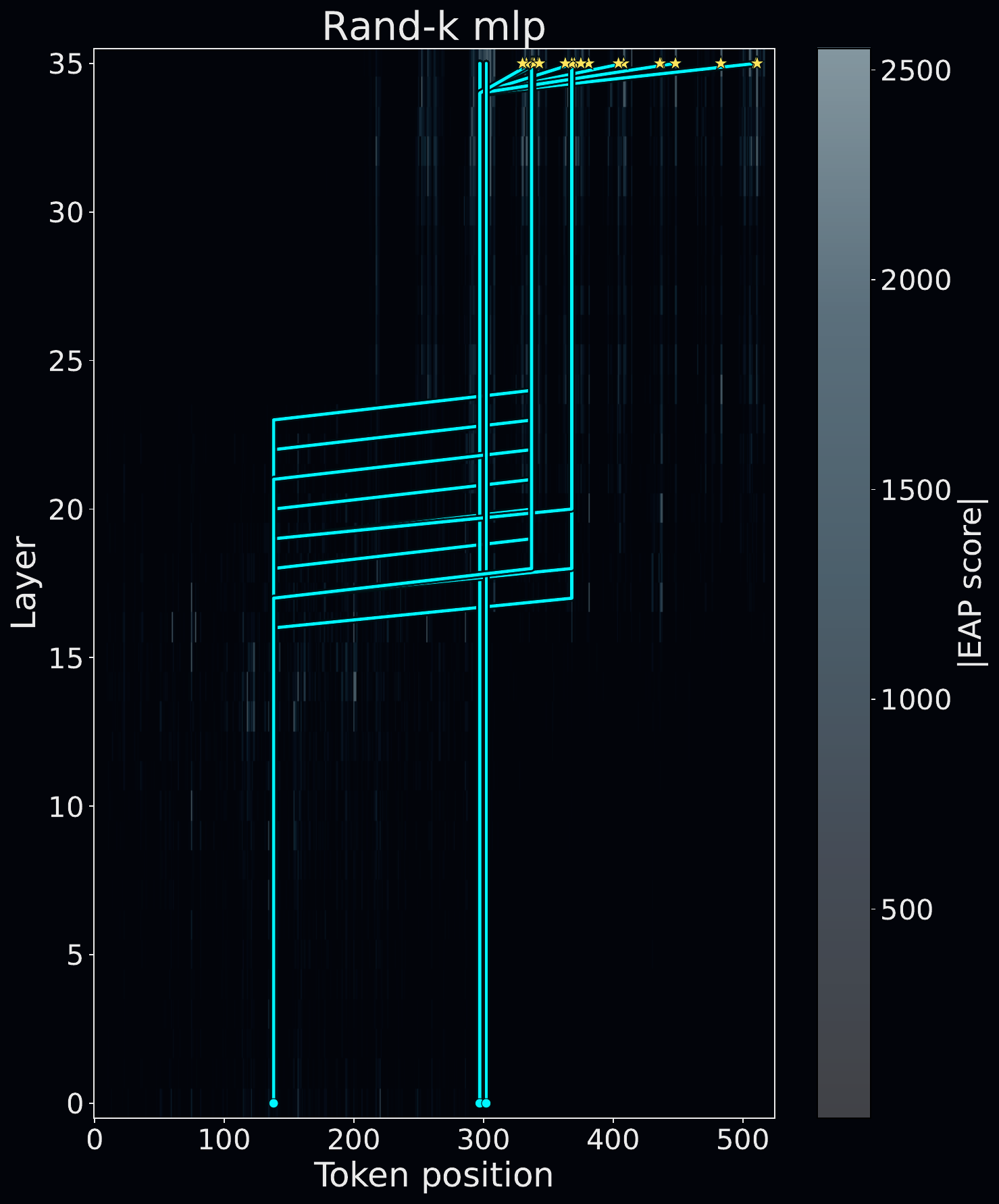}
\caption{Circuit heatmaps with boundary-path overlays for the edge-attribution patching diagnostic. Columns show coordinate-token Top-$k$, complement, and random-$k$ endpoint objectives; the top row shows attention heatmaps and the bottom row shows MLP heatmaps. The six panels use the individual heatmap PDFs for each objective and component type, preserving their titles, axes, overlays, and colorbars. The shared colorscale is anchored to the coordinate-token Top-$k$ condition, making the complement and random objectives visibly dimmer when their absolute edge scores are lower.}
\label{fig:eap-heatmaps}
\end{figure}

The heatmaps in Figure~\ref{fig:eap-heatmaps} add structural detail to the ablation result. All three attention circuits share their two strongest layers, layers $24$ and $34$, but the layer-$24$ dominance over layer $34$ is weaker for the coordinate-token Top-$k$ objective ($319{,}330/177{,}209=1.80\times$) than for the complement ($252{,}487/120{,}057=2.10\times$) and random-$k$ ($265{,}690/132{,}345=2.01\times$) objectives. The clearest separation appears in MLP edges: the coordinate-token Top-$k$ circuit peaks at layer $35$ ($262{,}061$) with layer $32$ second ($224{,}151$), whereas complement and random-$k$ objectives peak at layer $32$ and distribute more mass across layers $32$--$35$. The top visual source positions also shift: all objectives use positions $297$ and $302$, but the coordinate-token Top-$k$ circuit next emphasizes position $138$, whereas complement and random-$k$ objectives elevate neighboring positions $301$ and $303$. Endpoint answer positions similarly shift by one token at the boundary. This supports the precision/recall interpretation from Table~\ref{tab:eap-ablation}: coordinate-token-targeted edges concentrate near output-proximal coordinate-generation pathways, while detection-critical partial-occlusion behavior also uses broader late-layer routes.

\subsection{Quasi-layer RMSNorm control}
\label{app:rmsnorm-control}

Final RMSNorm can globally couple dimensions because each normalized coordinate depends on a shared denominator. Abstractly,
\begin{equation}
z_j
=
\gamma_j\frac{h_j}{R(h)},\qquad
R(h)
=
\sqrt{
\frac{1}{d}
\sum_{p=1}^{d}h_{p}^{2}
+
\epsilon
}.
\end{equation}
Here $p$ indexes hidden dimensions inside the RMSNorm denominator; $\ell$ is reserved for decoder-layer indices (Supp. Methods~\ref{app:notation-summary}). Changing one pre-norm coordinate can therefore rescale every post-norm coordinate. In principle, this shared denominator could convert a broad upstream magnitude change into a sharper final readout direction.

To test whether this coupling manufactures the endpoint/repair dissociation, we treat RMSNorm as an activation-conditioned quasi-layer and map post-norm directions back to the pre-norm residual stream by vector-Jacobian product. For a post-norm direction $g_{\mathrm{post}}$, the transported direction is
\begin{equation}
g_{\mathrm{pre}}
=
\mathcal{J}_{\mathrm{RMSNorm}}(h)^{\top}
 g_{\mathrm{post}},
\end{equation}
which can be computed without constructing the dense $d\times d$ Jacobian. For coordinate $j$,
\begin{equation}
g_{\mathrm{pre},j}
=
\frac{\gamma_j g_{\mathrm{post},j}}{R(h)}
-
\frac{h_j}{d R(h)^3}
\sum_{p=1}^{d} g_{\mathrm{post},p}\gamma_p h_p .
\end{equation}
We use this transported direction to compute pre-norm endpoint and repair rankings, then compare them with their post-norm counterparts. The control asks whether making final RMSNorm explicit changes which dimensions look repair-relevant or makes endpoint readout rankings more similar to corruption-derived repair rankings.

\begin{table}[t]
\centering
\caption{Supplementary detail for the quasi-layer RMSNorm ranking control using \texttt{bbox\_2d} readout rankings. The pre/post repair column compares repair-relevant rankings across the RMSNorm boundary. Endpoint-repair columns compare endpoint readout rankings with repair rankings before and after VJP transport. All values are means $\pm$ standard deviations across five bootstrap seeds.}
\label{tab:rmsnorm-app}
\scriptsize
\setlength{\tabcolsep}{4pt}
\resizebox{\textwidth}{!}{%
\begin{tabular}{lrrrr}
\toprule
$k$ & Pre/post repair overlap & Post endpoint-repair overlap & Pre endpoint-repair overlap & $\Delta$ pre-post \\
\midrule
$500$  & $\mathbf{.942\pm.008}$ & $.705\pm.013$ & $.713\pm.015$ & $\mathbf{+.008}$ \\
$1000$ & $\mathbf{.961\pm.004}$ & $.843\pm.008$ & $.843\pm.009$ & $\mathbf{+.001}$ \\
$1280$ & $\mathbf{.970\pm.003}$ & $.882\pm.006$ & $.877\pm.004$ & $\mathbf{-.004}$ \\
\bottomrule
\end{tabular}}
\end{table}

Table~\ref{tab:rmsnorm-app} shows that repair-relevant rankings are almost unchanged by crossing the final RMSNorm boundary: pre/post repair overlap is $.942\pm.008$ at $k=500$ and $.970\pm.003$ at $k=1280$. The VJP-transported endpoint ranking also does not close the endpoint-repair gap. The endpoint-repair overlap changes by only $+.008$ at $k=500$, $+.001$ at $k=1000$, and $-.004$ at $k=1280$. Thus, the dissociation is already present in the pre-norm residual representation and is not produced solely by the final shared normalizer.

A companion diagnostic using the original global \texttt{bbox\_2d} triage ranking gives repair-vs-readout overlaps of $.114$, $.252$, and $.344$ at $k=500$, $1000$, and $1280$, respectively. These values are reported separately because they compare the original field-token triage prior against repair dimensions, whereas Table~\ref{tab:rmsnorm-app} isolates the effect of transporting the same endpoint or repair signal through RMSNorm. The pre-norm intervention version of this control, in which selected pre-RMSNorm dimensions are patched and RMSNorm is recomputed downstream, is outside the reported scope; this result is therefore interpreted as a ranking-comparison control.
\subsection{Floored-clean endpoint recovery ablation}
\label{app:floored-clean-ablation}

As a supplementary-only ablation, we also evaluate a thresholded clean-replacement variant in which selected corrupted endpoint dimensions are replaced by the floored-clean value rather than the clean value:
\begin{equation}
h^{\mathrm{patch\mbox{-}floor}}_{t,I}=F_{\theta}(h^{\mathrm{clean}}_{t,I}),\qquad
h^{\mathrm{patch\mbox{-}floor}}_{t,\bar I}=h^{\mathrm{mask}}_{t,\bar I}.
\end{equation}
This ablation keeps the endpoint replacement set fixed while changing only the value inserted into the object-mask trace. It is not the primary recovery operator. The primary endpoint-recovery results use clean endpoint replacement as defined in Supp. Methods~\ref{app:dimension-sets} and~\ref{app:k-scaling}. The ablation is useful only for separating failures of index selection from failures caused by thresholding the inserted clean value.

\begin{table}[t]
\centering
\caption{Supplementary-only thresholded clean-replacement ablation for local-set endpoint recovery at $\theta=-0.5$. Entries report $R_{50}$; parentheses give the $z$ statistic against the matched random control. Short column headers follow the same index-set definitions as Tables~\ref{tab:local-set-recovery} and~\ref{tab:partial-occlusion}. These results are not the primary endpoint-recovery results in Table~\ref{tab:local-set-recovery}.}
\label{tab:thresholded-local-ablation}
\scriptsize
\setlength{\tabcolsep}{2.5pt}
\resizebox{\textwidth}{!}{%
\begin{tabular}{llrrrrrr}
\toprule
Ranking source & $k$ & Random & Fixed global & Split-local range & Sample-local readout & Sample-specific corruption & Bottom-ranked \\
\midrule
Coordinate-token & $500$  & $.1003\pm.0034$ & $.0943\,(-1.75)$ & $.0910\text{--}.0970\,(-2.74\text{--}-0.96)$ & $.0945\,(-1.71)$ & $.1021\,(+0.54)$ & $.0965\,(-1.13)$ \\
Coordinate-token & $1000$ & $.1734\pm.0046$ & $\mathbf{.1447\,(-6.27)}$ & $\mathbf{.1396\text{--}.1491\,(-7.38\text{--}-5.31)}$ & $.1456\,(-6.07)$ & $\mathbf{.2631\,(+19.65)}$ & $.1637\,(-2.11)$ \\
Coordinate-token & $1280$ & $.3196\pm.0748$ & $.2013\,(-1.58)$ & $.1967\text{--}.2076\,(-1.64\text{--}-1.50)$ & $.2963\,(-0.31)$ & $.3925\,(+0.98)$ & $.3441\,(+0.33)$ \\
\addlinespace[1mm]
Corruption-derived & $500$  & $.100\pm.0034$ & $.098\,(-0.69)$ & $.095\text{--}.099\,(-1.52\text{--}-0.44)$ & $.094\,(-1.71)$ & $.102\,(+0.53)$ & $.100\,(-0.04)$ \\
Corruption-derived & $1000$ & $.174\pm.0046$ & $.179\,(+1.20)$ & $.171\text{--}.176\,(-0.48\text{--}+0.49)$ & $.146\,(-6.07)$ & $\mathbf{.263\,(+19.65)}$ & $\mathbf{.191\,(+3.72)}$ \\
Corruption-derived & $1280$ & $.320\pm.0748$ & $.376\,(+0.75)$ & $.368\text{--}.393\,(+0.65\text{--}+0.98)$ & $.297\,(-0.31)$ & $.393\,(+0.98)$ & $.250\,(-0.93)$ \\
\bottomrule
\end{tabular}}
\end{table}

The ablation changes both the absolute recovery level and the relative ordering of several index sets. For the corruption-derived ranking family, every reported $R_{50}$ cell is lower under thresholded clean replacement than under the primary clean endpoint-replacement condition in Table~\ref{tab:local-set-recovery}. The effect is especially important at $k=1000$: the gap between the fixed population corruption-derived ranking and the sample-specific corruption-derived set is much larger under thresholded replacement, whereas clean endpoint replacement nearly closes it. The ablation therefore confirms that thresholding the inserted clean value can become the limiting factor and should not be conflated with the quality of the selected index set.

\subsection{Reporting and interpretation}
\label{app:reporting}

All hidden-state intervention studies report $R_{50}$, parseability where relevant, random-seed means, and size-matched controls. We interpret a dimension ranking as recovery-enriched only if its Top-$k$ sets outperform random and bottom-ranked controls at matched size and if withholding those dimensions disproportionately blocks endpoint recovery. We interpret an edge ranking as path-recovery-enriched only if ablating its top edges disrupts detection more than matched control edge rankings at the same edge budget. Otherwise, the ranking is treated as readout-informative but recovery-incomplete.

This reporting convention preserves the central distinction of the study: endpoint rankings can be useful references for circuit tracing without being sufficient explanations of visual-corruption repair.

\end{document}